# Joint Modeling and Registration of Cell Populations in Cohorts of High-Dimensional Flow Cytometric Data

Saumyadipta Pyne[1,2], Kui Wang[3], Jonathan Irish[4,5,6], Pablo Tamayo[2], Marc-Danie Nazaire[2], Tarn Duong[7], Sharon Lee[3], Shu-Kay Ng[8], David Hafler[2,9], Ronald Levy[4], Garry Nolan[5], Jill Mesirov[2], and Geoffrey J. McLachlan[3]

[1]CR Rao Advanced Institute of Mathematics, Statistics and Computer Science, Hyderabad, India [2]Broad Institute of MIT and Harvard, 7 Cambridge Center, Cambridge, MA 02142, USA. [3]Department of Mathematics, University of Queensland, St. Lucia, Queensland, 4072, Australia. [4]Division of Oncology, Stanford Medical School, Stanford, CA 94305, USA [5]Baxter Laboratory for Stem Cell Biology, Department of Microbiology and Immunology, Stanford School of Medicine, Stanford, CA 94305, USA. [6]Department of Cancer Biology, Vanderbilt University, Nashville, TN 37220, USA. [7]Unit Mixte de Recherche, 144/Centre National de la Recherche Scientifique, Institut Curie, Molecular mechanisms of intracellular transport Laboratory, Paris, France. [8]chool of Medicine, Griffith University, Meadowbrook, QLD 4131, Australia. [9]Department of Neurology, Yale School of Medicine, 15 York Street, New Haven, CT 06520, USA.

**Abstract**

In systems biomedicine, an experimenter encounters different potential sources of variation in data such as individual samples, multiple experimental conditions, and multivariable network-level responses. In multiparametric cytometry, which is often used for analyzing patient samples, such issues are critical. While computational methods can identify cell populations in individual samples, without the ability to automatically match them across samples, it is difficult to compare and characterize the populations in typical experiments, such as those responding to various stimulations or distinctive of particular patients or time-points, especially when there are many samples. Joint Clustering and Matching (JCM) is a multi-level framework for simultaneous modeling and registration of populations across a cohort. JCM models every population with a robust multivariate probability distribution. Simultaneously, JCM fits a random-effects model to construct an overall batch template – used for registering populations across samples, and classifying new samples. By tackling systems-level variation, JCM supports practical biomedical applications involving large cohorts.

## 1 Introduction

Flow cytometry is widely used for single cell interrogation of surface and intracellular protein expression by measuring fluorescence intensity of fluorophore-conjugated reagents. Recent technical advances have taken the field towards single cell proteomics[1] and enabled highly multiparametric analysis[2] and computational cytomics[3]. Consequently, applications in systems biomedicine are presenting new challenges to cytometric analysis. Increasingly such studies involve cohorts with large numbers of patients, replicates, and may also use multiplexing of marker staining panels for probing large signaling networks[4]. Further, the development of mass cytometry promises the ability to compare 50-100 features per cell[5, 6]. Owing to multiple reasons such as variation among individuals in a cohort, simultaneous use of different stimulation conditions and panels in a given experiment, biological and technical replicates, the highly multivariate nature of the new platforms' measurements, etc., the resulting datasets are rich and complex. Currently there exists no single standard procedure for performing reproducible cohort-wide analysis while tackling systems-level heterogeneity and noise in multiple samples.

Recently, we developed a platform (FLAME) for automated analysis of high-dimensional flow data[7]. Each cell population (henceforth simply called population) in a sample is modeled by FLAME as a cluster of points with similar fluorescence intensities in the multi-dimensional space of markers. FLAME's heavy-tailed and asymmetric distributions are especially appropriate for flow data, since rare and interesting subpopulations tend to be tails connected to larger populations[8]. Notably, the field of computational cytomics has witnessed rapid growth in the past few years, as reviewed by Lugli et al[3].

While modeling populations in flow data remains a difficult problem, a second and even more important challenge appears when there are many samples and conditions to compare – how to efficiently match or "register" the corresponding populations across a *batch* of samples. The difficulty of this problem arises from (a) the high-dimensionality of data, which prevents visual matching of populations, (b) large cohort or batch sizes, and (c) high inter-sample variation, all of which make the manual approach challenging. Yet it is essential to determine the batch-wise correspondence among populations with automation so that we can register them in high-dimension, which enables direct quantitative comparison of samples across conditions, phenotypes or time points. Addressed with algorithmic precision and rigor, automatic registration can facilitate clinical applications with diagnostic or prognostic implications. For instance, it can be useful for monitoring of specific cellular events such as lymphocytic infiltration in tumors, immuno-profiling of patients following treatment, etc.[9, 10]. By creating parametric models of the matched spatio-temporal profiles, we can use the estimated model parameters to accurately classify new samples as well as identify aberrant patterns (outliers).

A composite solution to these two complex problems – modeling each population within a sample, and registering them across samples – marks a significant improvement over FLAME and the other predominantly clustering approaches[3]. Currently, FLAME first models the populations separately within individual samples, and then tries to match these populations post hoc by running an external module (using Partitioning Around Medoids or PAM) on the model parameters. We observed that this approach has several limitations. For instance, meta-clustering can be overly sensitive to the accuracy of the comparison results of PAM. Indeed it is difficult to tackle high inter-sample variation in a batch via post hoc comparison of any particular parameter. Further, while meta-clustering only matches pairs of certain population-features (e.g. locations), much more robust and overarching relationships among the same populations can be captured with a higher batch-level model. Finally, we note that without modeling the whole batch simultaneously, an overall consensus template of the batch cannot be formed. In that sense, FLAME and other algorithms that analyze single samples cannot determine batch characteristics systematically.

We present a new multi-level framework called Joint Clustering and Matching (JCM) that operates on an entire batch of samples across 2 levels: (1) at a sample-specific "lower" level, JCM models every cell population as a cluster (i.e. a *component* of a finite mixture model of multivariate $t$ or skew $t$-distributions); and simultaneously, (2) at a batch-specific "higher" level, JCM constructs a parametric *template*, which models overall characteristics of a batch. JCM achieves this by fitting a Random-Effects Model (REM) that allows every sample in a given batch to be modeled as an instance of an "original" template possibly transformed with a flexible amount of variation. Here we also describe our Expectation–Maximization (EM) algorithm for efficient model fitting. Its multi-level design gives JCM the ability to establish a direct parametric correspondence between each cell population in the batch template and its counterpart within an individual sample. Unlike FLAME, this allows JCM to explicitly tackle inter-sample variation, a common concern for flow data, and thus support both biological and clinical applications.

In recent years, researchers have also started multiplexing many staining panels to overcome limits on the numbers of markers that can be accurately measured together using commercial cytometers[4]. While the resulting data are more enriched, it can also produce a large number of distinct features from every panel of markers. Currently there exists no technique for systematic integration of such features across panels into meta-features for the common underlying sample. As part of JCM analysis, we introduce a new technique to combine both univariate and multivariate JCM features across multiplexed panels to construct enriched meta-features (or *feature-sets*), and use these to improve sample classification.

Using simulation as well as several real-world benchmark datasets, we found that key performance attributes such as classification accuracy and running time of JCM are quite favorable compared to other methods. Further, we applied JCM to two cell signaling datasets. First, we used it to obtain multi-parametric characterization of different T cell subpopulations upon T cell receptor (TCR) stimulation in a time course phosphorylation experiment. This illustrates how a complex multi-class and multi-sample experiment can be systematically analyzed in a fully automated and reproducible manner to generate precise and objective profiles for every class. Importantly, it is based on a comprehensive list of rigorously estimated model parameters for each population, which is output by JCM. As illustrated by our next application, such unsupervised, thorough approach can also reveal new or subtle expression phenotypes in specific subpopulations, which might otherwise go undetected in manual gating that generally follows a pre-determined sequence of 2D visualization steps. We applied JCM to understand differential patterns of altered B cell receptor (BCR) signaling in human follicular lymphoma (FL) tumor samples. By combining JCM features from multiplexed panels of 16 phospho-markers, we identified a novel spatio-temporal signature of BCR signaling in a specific subpopulation of the lymphoma B cells that improved the separation between 2 classes of patients previously reported by Irish et al.[9] to have markedly different survival. We also devised new visual means for overlaying expression templates to capture the variation in data both within and across a batch. This highlights the capability of JCM to distinguish complex biological contexts via quantitive class-specific characteristics, which may be very useful in new studies involving large cytometric cohorts.

# 2 Results

## 2.1 Overview of JCM

JCM is run in the following sequence of steps (flowchart in Supplementary Fig. S1) –

**(1)** Obtain the expression matrices from an input batch of preprocessed samples.

**(2)** Fit a 2-level model (as illustrated in Fig. 1) to these data such that —

**(2a)** an overall parametric template for the batch is constructed by modeling the affine transformations that may exist among the corresponding populations across samples, and simultaneously

**(2b)** every sample is modeled with its own mixture of skewed and heavy-tailed multivariate probability distributions, which characterizes the high-dimensional populations while registering them using the batch template.

**(3)** Output files are produced containing the fitted models for the batch template and all samples – in formats suitable for visualization and downstream analysis programs. Novel overlay plots are produced for visual comparison of all class-templates.

To illustrate the different capabilities of JCM, we applied it to two sets of experiments involving multiple markers, time points (or stimulations), staining panels, and sample classes. Also, we allow two modeling options: the default using mixtures of multivariate skew $t$-distributions and its symmetric counterpart using a mixture of multivariate $t$-distributions, and tested both models on two cell signaling datasets.

## 2.2 Spatio-temporal characterization of TCR activation

We analyzed phosphorylation patterns downstream of T cell receptor (TCR) activation in naïve and memory T cells across six classes of samples corresponding to six time points: 0, 1, 3, 5, 15, and 30 min originally measured by Maier et al[11]. In that study, human expertise played a key role in manually and visually identifying each population in every sample at every time-point, and then carefully comparing them based on selected features of chosen populations. In the process, many manual decisions were taken and highly supervised time-consuming operations were performed repeatedly such as the applied sequence of gates, the selection of useful parameters for comparing the subsets across classes, etc. Traditionally, therefore, the results of manual gating even on similar experiments can vary with such decisions, which in turn depend on the experience of the human expert.

JCM, in contrast, produced the full sequence of spatio-temporal expression phenotypes of phosphorylation in five distinct subsets of T cells, which are matched across all samples. These 5 populations were characterized in a fully unsupervised manner in 4-dimensional marker-space, as well as in terms of the $5^{th}$ dimension of time. The model yielded a comprehensive list of matched high-dimensional parameters, not just a few pre-determined visual (i.e. 2-D) features. This list could be readily used for exploratory statistical analyses (e.g. feature selection, discriminant analysis) to accurately identify the changes in every population over time. Since the cohort was modeled as a batch by JCM, we can also compare the overall batch-templates computed for every time-point, both statistically and visually, to cature the longitudinal phenotypic trend starting from the activation of TCR up to its de-activation. Thus the JCM framework is objective, fast, quantitive and reproducible.

The sequence starts at 0 min, prior to stimulation with an anti-CD3 antibody (baseline measurement), reached peak levels of phosphorylation at 3-5 min. and then subsided by 30 min. JCM's multi-level modeling of the time course data is illustrated in Fig. 1(a). The profile of each of the 5 populations (denoted #1–5) were distinguished apart, matched across samples, summarized with templates and compared across 6 time-points. The overall changes summarized as high-dimensional templates for each of the successive classes can be observed in the Supplementary Fig. S2. The overall spatio-temporal differences both within and across classes may be observed with JCM's new overlay plots (Supplementary Fig. S3). Specifically, the alterations in the naïve and memory T cell populations are outlined in Supplementary Fig. S4. For details on the experiments, see Supplementary Information.

Two markers in the staining panel, CD4 and CD45RA, were used for characterizing the different populations, while two other markers, SLP76 (p-Y128) and ZAP70 (p-Y292), were used to measure the intensity of phosphorylation in these subsets. As described in Maier et al.[1], we used the signatures $CD4^{hi}$ with $CD45RA^{hi}$ and $CD4^{hi}$ with $CD45RA^{lo}$ to represent the primarily naïve and memory T cell subsets, respectively. Upon fitting JCM-MVT model to each of the 6 classes, an overall pattern for 5 matched populations emerged (indexed #1 through #5 in Fig. S2(a)-(e)). As expected, a rapid rise in the intensities of phosphorylation markers SLP76 and ZAP70, especially the latter, was observed soon after stimulation for all populations with the possible exception of #2. While both naïve (#3) and memory T cell subsets (#4) showed similar peak levels of phosphorylation initially (Fig. S2(c)-(d)), the former exhibited a faster decline with time (Fig. S2(d)-(e)), consistent with prior results[1]. In fact, both $CD45RA^{+}$ populations (#1 and #3) exhibited similar expression throughout. Upon p-CD3 (p-Y142) normalization, higher phosphorylation in memory T cells compared to naïve T cells between 5 and 15 min – as observed manually[1] – was recapitulated with help of JCM.

## 2.3 BCR signaling feature-sets distinguish FL subclasses

In a recent study based on human expert analysis, Irish et al.[9] stratified follicular lymphoma (FL) patients into 2 classes with markedly different overall survival depending on the presence or absence of a Lymphoma Negative Prognostic (LNP) subset of B cells in tumor. The LNP cells showed altered BCR signaling, and were identified by the expressions of a multiplexed panel of selected phospho-markers. The signaling based stratification of patients into $LNP^{+}$ and $LNP^{lo}$ classes is therefore of clinical significance. We used JCM for (a) automation — to systematically combine features from multi-panel data from FL patients, and (b) discrimination — to identify features that could separate the pre-defined FL patient classes as best as possible.

Through automated analysis of multiplexed data, JCM identified a nuanced signature for signaling alterations in high-dimensional marker-space that further improved the distinction between the two FL patient classes. We analyzed 28 pre-processed patient samples for two time points, 0 min and 4 min (i.e. pre- and post-BCR stimulation respectively). At every time-point, and for all patients, the data consisted of 8 panels, each with 4 markers, including two B cell markers CD20 and BCL2 that were common to every panel. Signaling responses were measured in terms of phosphorylation of 16 phospho-proteins from the BCR signaling network. By multiplexing panels, the signaling for all these network components could be measured in every sample. Each sample's phenotype (or class label), $LNP^{lo}$ (18 samples) or $LNP^{+}$ (10 Samples), was assigned by human expert analysis (Supplemental Methods of Irish et al.[9]).

For both unstimulated (0 min) and stimulated (4 min) conditions, each class of patient samples were modeled with JCM-MST using two-component multivariate skew $t$ mixture models. The templates revealed the class-specific features of two lymphoma B cell populations. For convenience, let us call these two populations "mound" and "base" corresponding to higher and lower levels of stimulation respectively. These are components of the JCM mixture model that primarily represent populations in which BCR signaling is intact (i.e. non-LNP cells) as opposed to altered (LNP cells). The change between the corresponding features pre- and post-stimulation provided a kind of baseline correction to the resting level of signaling for each sample. This approach corresponds to asking whether the response of lymphoma B cells to BCR engagement was heterogeneous, but using the entire set of continuous features for exploring tumor heterogeneity rather than only median phosphorylation, the primary discretized feature in the Irish et al. study[9].

We introduced a new strategy for a combined analysis of multiplexed markers probing different parts of the BCR signaling network. The JCM features of 16 phospho-markers across all 8 panels were pooled for identifying enhanced meta-features (or feature-sets analogous to the concept of gene-sets). Thus we applied Gene Set Enrichment Analysis (GSEA[12]) to every feature-set to test their abilities to distinguish between $\mathrm{LNP}^{\mathrm{lo}}$ and $\mathrm{LNP}^{+}$ samples. Notably, Irish et al.[9] had previously discovered that the size of the LNP population could be used to distinguish FL patients into 2 classes with different outcomes. However, these results were based on manual demarcation of the LNP subset, and therefore based on low-dimensional gating of data. Interestingly, in our feature-set enrichment analysis, the single most significantly enriched feature-set (at $P$-value level 0.05 by Kolmogorov-Smirnov test of GSEA[12]), i.e. the most distinctive meta-feature across these 2 patient classes, was skewness ($\delta$) of the mound at 5 min. ($P$-value 0.0144, $q$-value 0.058; Supplementary Fig. S5). Across $\mathrm{LNP}^{\mathrm{lo}}$ and $\mathrm{LNP}^{+}$ classes, this spatial signature (i.e. stimulated mound skew) is distinctive both visually (Fig. 2(a) and 2(b)) and statistically (average posterior log-odds ratios[13] in Fig. 2(c)), particularly for markers such as p-PLCg2, p-BLNK, and p-SFK (Supplementary Fig. S6). In particular, we draw attention to Fig. 2(a), outlining the asymmetric expression of the mound in $\mathrm{LNP}^{\mathrm{lo}}$ samples, which contrasts with their more spherical counterparts (i.e. lower skew) in the $LNP^{+}$ samples. The distinction is in fact statistically significant even after controlling for the corresponding base (LNP) $\mathrm{LNP}^{+}$ population sizes (e.g. for p-SFK the GLM based p-value after controlling is 0.0079).

The skewness, given by the parameter vector $\boldsymbol{\delta}$, of the stimulated mound in $\mathrm{LNP}^{+}$ samples is expressed in form of a heavy left tail (Fig. 2(b)). This suggests the likely presence of a subpopulation of primarily non-LNP cells with partially altered signaling at a given time-point. Whether it is of real prognostic value needs to be tested in future studies. Our main point is that JCM's automatic feature detection can reveal new spatio-temporal states and their characteristics. State transitions can be numerically measured and monitored even if they are subtle across classes. For instance, if the alteration in BCR signaling is gradual and not sharp, then it can be difficult to demarcate or determine the size of the LNP component accurately, and yet the skew feature can be used for nuanced understanding of the same population thus providing mechanistic insights into the biology of the system in action.

## 2.4 Performance and Comparactive Analysis

Although in general flow analysis methods do not compute cluster correspondence, we compared JCM with FLAME[7] and pooled metaclustering, and another published method flowClust[14]. Based on four real-world benchmark datasets from flowCAP1 contest (N. Agheepour et al. in preparation), we first created templates for each panel and class. Then every sample was classified with the most similar template using the empirical (sample) version of the Kullback-Leibler distance. A comparison of the classification error rates for all methods (see Supplementary Table T1) demonstrates the effectiveness of the JCM approach. The JCM approach is based on the assumption that a sample can be modeled as an instance of its parent template, and hence the accuracy of the JCM approach for classification purposes depends on how distinctive the class templates are. In the iterative fitting of a class template via the EM algorithm, careful consideration needs to be given to the choice of starting values since the likelihood function will usually have multiple local maxima. Our choice of starting values included initializing the iterative process via a rapid, robust $k$-means based clustering using flowMeans[15].

Using random-effects modeling within the mixture modeling framework did not affect JCM's computational performance (Supplementary Figure S7). Simulation shows that the running time per EM loop is typically below 1 minute on a standard quad-Core desktop PC, and is linearly proportional to the number of samples, the number of points per sample, and the number of clusters. Therefore, JCM is scalable to analyze data from fairly large cohorts.

# 3 Discussion

High-dimensional computational analysis of flow data is receiving increasing attention with the rapid rise in the number of markers that can be used to probe each cell in parallel[3, 6]. Mirroring the perception of a flow sample as a mixture of cell populations, finite mixture of Gaussians has long been an attractive modeling mechanism[16]. Recently, robust mixture models with multivariate $t$ and skew $t$ distributions were introduced for analyzing flow data with non-Gaussian features such as outliers, heavy tail densities, and asymmetric shapes[7, 14, 17]. In addition to modeling cell populations, Pyne et al.[7] also highlighted the importance of registering them across samples for the purpose of automated analysis of classes and conditions. For re-structuring of cell populations, recent studies have noted that the optimal algorithmic strategy to do so is in conjunction with data modeling[14, 18].

The key contribution of JCM is its joint approach to address two challenges with a single composite model. It is a two-level framework for simultaneous mixture modeling and registration of populations in an entire batch of flow samples. It allows JCM to meet a key need of cytomics – reproducible analysis of data from many samples and conditions – by matching the populations internally and simultaneously. Notably, in the field of pattern recognition, alignment of images and curves in lower-dimensional space have emerged as active areas of research in recent years[19, 20, 21]. Thus, JCM provides an important extension to algorithms like Gaussian mixture regression models[21] to multivariate $t$ and skew $t$-distributions, which are fit with the EM algorithm. This algorithm is an effective generic technique for parameter estimation[22], and we introduce new algorithmic procedures for the JCM-specific application of EM (Appendix A, Supplementary Information).

Automated population registration of JCM marks a significant technical improvement over FLAME (Performance and Comparative Analysis, Supplementary Information). Unlike the post-hoc meta-clustering program of FLAME, matching of populations by JCM is intrinsic to its modeling strategy. It is achieved by fitting a random-effects model (REM), a standard meta-analytic approach for estimating the mean of a distribution of effects[23]. Rare past usage of REM in cytomics was limited to measuring variability of very specific features, e.g., CD4 expression[24]. JCM is perhaps the first framework that incorporates REM for comprehensive batch characterization in flow data analysis (Fig. 1). In particular, REM used affine transformation parameters to explicitly learn relationships among every population in a batch even in the presence of flexible amounts of cross-sample variation. We are unaware of the existence of any method other than JCM that computes batch characteristics of flow samples. If JCM were to be reduced to its lower level, that is, clustering only, and further restricted to a single sample input, then it would be similar to FLAME clustering. The latter was nonetheless ranked by rigorous benchmarking and expert analysis to be among the top performing unsupervised algorithms at a recent international contest on flow analysis FlowCAP1 organized in NIH [31]. This underscores much greater potential of JCM with its more flexible approach than FLAME as noted above.

A technical advantage of JCM's REM-based registration is that it accounts for the populations' scaling and shifting transformations without explicitly "correcting" them. Aligning populations can be useful at the preprocessing stage for application of common gates or filters en bloc. However, for precise modeling of populations, we want to identify the spatio-temporally distinctive high-dimensional features, which may be characteristic of each sample's multi-marker phenotype. While we do not want to homogenize features by aligning them, at the same time, we do want to register populations as they appear in high-dimensional space with precision and rigor. This makes registration more challenging than matching (as in FLAME meta-clustering[7]) or alignment (as in channel normalization[25]). In fact, we compared JCM with FLAME meta-clustering on benchmark data, and as shown in Appendix B, Supplementary Information, JCM templates keep classification error rates low in the face of increasing inter-sample variation in batches derived from real cytometric cohorts.

Perhaps the most attractive feature of REM is an overall consensus template that emerges connecting both levels of JCM output. Thus JCM can establish a direct parametric correspondence between each population in the batch template and its counterpart within every sample. Further, the template allows JCM to capture cross-sample inter-relationships that may exist among populations and are useful for accurate registration. For instance, if a certain population A generally appeared in between two populations B and C, then it is useful to learn about such relative positioning of A even if its absolute location varied from sample to sample. It makes JCM more robust to common transformations (such as shifting or scaling of populations – to which these relationships are generally invariant) compared to FLAME meta-clustering, which can handle only limited variation in absolute locations. Thus the JCM template provides a "ground truth" while the REM transformation parameters quantify each individual instance's deviation from that reference structure. From practical standpoint, given that the JCM templates are defined by parametric distributions, they allow direct statistical comparison of batches which could represent, say, different subclasses of patients or successive longitudinal observations. We devised novel overlay plots for visual comparison of overall batch-structures along every dimension both within and across classes. Moreover, any new patient sample could be easily classified with the group that has the most similar template (as determined by, say, Kullback-Leibler distance). Finally, a JCM template can offer the user a convenient and rigorous summary of a given cohort's overall population structure.

Parametric characterization of cohorts in terms of their high-dimensional spatio-temporal features can reveal complex and dynamic biological contexts and present them for further investigation. Dissecting and monitoring the parameters of individual cellular species as they evolve over time — such as our time course profiling of TCR stimulation (Fig. 2) — could be useful in many biomedical applications. The JCM models supporting asymmetric and heavy-tailed distributions of events are uniquely suited for detecting features that appear dynamically as hard-to-separate transitional features, such as asymmetric or tail subpopulations[8], that are otherwise difficult to distinguish via automation. Further, by pooling features across multiplexed staining panels, JCM can detect complex biological contexts involving multiple markers from a signaling pathway or network[9].

JCM can serve as a practical framework that is suitable for clinical applications. Here, its main objective is to learn the specific target populations' parameters for large numbers of samples precisely and quickly. Yet, in Clinical mode, the modeling must also be robust enough to allow a reliable parameter-driven classification of patient samples. This is of particular concern for flow data which may contain high inter-sample variation due to the presence of complex, biologically interesting subpopulations, along with noise, within the target pool of primary cells. Explicit detection of variation by REM is useful for batch characterization, QA/QC, as well as downstream analysis. Moreover, JCM produces an array of new, insightful plots. For instance, the overlay plot can reveal within-class variation along any dimension, while the intensity heatplots take advantage of REM to allow monitoring of spatio-temopral changes in individual populations that are matched across the cohort. Another attractive practical feature of JCM is its representation of output in the form of a generic feature-by-sample matrix, which can analyzed with common bioinformatic pipelines. Thus, here we used the well-known GSEA algorithm[12] to create a new technique for combining JCM features into enriched meta-features across multiplexed staining panels. The simple new technique may become highly effective as more multiplexed staining data begin to appear[6].

The random-effects model, by accounting for sample-specific variation provides an intrinsic cohort-wide meta-analysis. As a multi-level model, JCM design can be generalized further to include even higher level parameterization for representing time points or patient subtype information (e.g. clinico-pathological phenotypes or genotypes). This makes JCM well suited for integrative cytomics. In fact, our simulations show that besides being efficient in batch mode analysis, JCM is also robust against both class-size and the amount of inter-sample variation it can handle (Supplementary Information). For instance, the running time for JCM modeling of a sample in our phosphorylation data averaged 33.7 sec per sample on a standard quad-core desktop PC. This contrasts sharply with the hours of manual analysis performed over weeks by multiple researchers in the original study. With increasing multi-parameterization and multiplexing of cytomic data, JCM can facilitate automated, quantitative, scalable and objective investigation of complex hypotheses about different conditions and cohorts of biomedical interest.

# 4 Methods

Following is the description of the JCM workflow and details of the models and methods, also continued in Supplementary Information.

**Step 1: JCM input:** The input to JCM is a batch of $m$ flow cytometric samples in the form of either a zipped folder of $m$ .fcs files or a R flowSet object of $m$ flowFrames. We assume that the flow data have been acquired, quality-controlled and preprocessed (such as live cell gating) properly. Commercial software and freely available BioConductor packages (e.g. flowCore[26]) are highly useful for such purposes. If $q$ (>1) multiplexed panels of markers were used, then $q$ such zipped folders or flowSet objects matrices must be provided, such that every sample is represented by $q$ panels. The user can also specify which mixture model (MVT or MST) to fit, and an optional range for the expected number of populations ($g$) in the batch template.

For each sample $k$, we extract from its .fcs file or R flowFrame an $n_k$ by $p$ expression matrix, corresponding to fluorescence intensity values of $p$ markers or antibodies for $n_k$ cells. (Typically, $p$ varies between 4 and 8 but could be as high as 17 in fluorescence cytometry and 35-40 in mass cytometry, but JCM is not limited by any particular value of $p$; $n_k$ could range from hundreds to hundreds of thousands per sample; $q$ is currently a moderate constant such as 10 or less whereas $m$ could be in hundreds.)

**Step 2: Multi-level modelling:** A two-level model is fitted to an input batch or class $C$ of $m$ samples where each sample is represented by its own $n_k \times p$ expression matrix, where $k$ indexes the sample ($k = 1, \ldots, m$). The problem is to simultaneously (a) model all $m$ samples in a batch while (b) creating a $p$-dimensional template of $g$ components for matching the corresponding populations across all samples. Below we describe the JCM model, for both symmetric and asymmetric components, which are fitted with the JCM-specific EM algorithm for maximum likelihood (ML) estimation as described in detail in Appendix A (Supplementary Information).

Let $\boldsymbol{y}$ denote a $p$-dimensional vector denoting the values of the $p$ markers in a sample. Then JCM provides a method for constructing a template density of $y$ for a class of $m$ samples, where we let $\boldsymbol{y}_k$ denote the data observed in the $k$th sample ($k = 1, \ldots, m$). For the construction of the template density, we use a mixture of $g$ component distributions, where the latter are members of the $t$-family of distributions[27] or of a skew-extension of this family[7]. In order to define these component distributions, we consider first the $g$-component normal mixture density, which can be expressed as

$$f(\boldsymbol{y}; \boldsymbol{\Psi}) = \sum_{h=1}^{g} \pi_h f(\boldsymbol{y}; \boldsymbol{\theta}_h), \tag{1}$$

where $f(\boldsymbol{y}; \boldsymbol{\theta}_h) = \phi(\boldsymbol{y}; \boldsymbol{\mu}_h, \boldsymbol{\Sigma}_h)$ and $\phi(\boldsymbol{y}; \boldsymbol{\mu}_h, \boldsymbol{\Sigma}_h)$ denotes the $p$-variate normal density with mean $\boldsymbol{\mu}_h$ and covariance matrix $\boldsymbol{\Sigma}_h\,(h = 1, \ldots, g)$; $\pi_h, \ldots, \pi_g$ denote the mixing proportions which are non-negative and sum to one. The vector $\boldsymbol{\theta}_h$ denotes the elements of $\boldsymbol{\mu}_h$ and the elements of $\boldsymbol{\Sigma}_h$ known a priori to be distinct. The vector of unknown parameters is given by $\Psi = (\pi_h, \ldots, \pi_{g-1}, \boldsymbol{\theta}_1^T, \ldots, \boldsymbol{\theta}_g^T)$, where the superscript $T$ denotes vector transpose. In (1), $f$ is being used generically to denote a density function.

In the present context where the tails of the normal distribution are heavier or the parameter estimates are affected by atypical observations (outliers), the fitting of mixtures of multivariate $t$-distributions provides a more robust approach to the fitting of normal mixture models[27]. The $t$-component density with location parameter $\boldsymbol{\mu}_h$, positive-definite scale matrix $\boldsymbol{\Sigma}_h$, and $\nu_h$ degrees of freedom is given by

$$t_p(\boldsymbol{y}; \boldsymbol{\mu}_h, \boldsymbol{\Sigma}_h, \nu_h) = \frac{\Gamma\left(\frac{\nu_h+p}{2}\right)|\boldsymbol{\Sigma}_h|^{-1/2}}{(\pi\nu_h)^{p/2}\Gamma(\nu_h/2)\{1 + d_h(\boldsymbol{y})/\nu_h\}^{(\nu_h+p)/2}}, \tag{2}$$

where $d_h(\boldsymbol{y}) = (\boldsymbol{y} - \boldsymbol{\mu}_h)^T \boldsymbol{\Sigma}_h^{-1}(\boldsymbol{y} - \boldsymbol{\mu}_h)$ denotes the Mahalanobis squared distance between $\boldsymbol{y}$ and $\boldsymbol{\mu}_h$ (with $\boldsymbol{\Sigma}_h$ as the scale matrix), and $\Gamma(\cdot)$ denotes the Gamma function. The parameter $\nu_h$ acts as a robustness tuning parameter, which can be inferred from the data by computing its maximum likelihood estimate.

In order to reliably model the clusters that are not elliptically symmetric but are skewed, we shall adopt component densities that are a skewed version of the $t$-distribution. Over the years, a number of proposals have been put forward with increasing level of generality for a skew form of the $t$-distribution. We shall adopt the version proposed by Sahu et al.[28], which is quite general. Accordingly, we let $\boldsymbol{\Delta}_h$ be a diagonal matrix with diagonal elements given by the vector $\boldsymbol{\delta}_h = (\delta_{1h}, \ldots, \delta_{ph})^T$ of skewness parameters. Suppose that conditional on $w$ and membership of the $h$th component,

$$\begin{pmatrix} \boldsymbol{U}_0 \\ \boldsymbol{U} \end{pmatrix} \sim N\left(\begin{pmatrix} \boldsymbol{\mu}_h \\ \mathbf{0} \end{pmatrix}, \begin{pmatrix} \boldsymbol{\Sigma}_h/w & \mathbf{O}_p \\ \mathbf{O}_p & \boldsymbol{I}_p/w \end{pmatrix}\right), \tag{3}$$

where the random variable corresponding to $w$ is distributed according to the gamma$(\nu_h/2, \nu_h/2)$ distrubution. In the above, $|\boldsymbol{U}|$ denotes the vector whose $i$th element is equal to the magnitude of the $i$th element of the vector $\boldsymbol{U}$, $\mathbf{0}$ denotes the $p$-dimensional null vector, $\mathbf{O}_p$ denotes the $p \times p$ null matrix, and $\boldsymbol{I}_p$ denotes the $p \times p$ identity matrix.

Then

$$\boldsymbol{Y} = \boldsymbol{\Delta}_h|\boldsymbol{U}| + \boldsymbol{U}_0 \tag{4}$$

defines a $p$-dimensional multivariate skew $t$-distribution with location $\boldsymbol{\mu}_h$, scale matrix $\boldsymbol{\Sigma}_h$, skew (diagonal) matrix $\boldsymbol{\Delta}_h$, and $\nu_h$ degrees of freedom. Its density can be expressed as

$$f(\boldsymbol{y}; \boldsymbol{\mu}_h, \boldsymbol{\Sigma}_h, \boldsymbol{\Delta}_h, \nu_h) = 2^p t_p(\boldsymbol{y}; \boldsymbol{\mu}_h, \boldsymbol{\Omega}_h, \nu_h) T_p(\boldsymbol{y}^*; \boldsymbol{\mu}_h, \boldsymbol{\Omega}_h, \nu_h + p), \tag{5}$$

where $\boldsymbol{\Omega}_h = \boldsymbol{\Sigma}_h + \boldsymbol{\Delta}_h \boldsymbol{\Delta}_h^T$, $\boldsymbol{\Lambda}_h = \boldsymbol{I}_p - \boldsymbol{\Delta}_h^T \boldsymbol{\Omega}_h^{-1} \boldsymbol{\Delta}_h$, $\boldsymbol{y}^* = [(\nu_h + p)/\{\nu_h + d_h(\boldsymbol{y})\}]^{1/2} \boldsymbol{\Delta}_h^T \boldsymbol{\Omega}_h^{-1} (\boldsymbol{y} - \boldsymbol{\mu}_h)$. In (5), $t_p(\boldsymbol{y}; \boldsymbol{\mu}_h, \boldsymbol{\Sigma}_h, \nu_h)$ denotes the $p$-variate $t$-density with location $\boldsymbol{\mu}_h$, scale matrix $\boldsymbol{\Omega}_h$, and degrees of freedom $\nu_h$, and $T_p$ denotes its ($p$-variate) distribution function.

We represented the class template by fitting the $g$-component mixture model in (1) to all the $m$ samples considered simultaneously, using (2) to represent the $t$-component densities in the symmetric case and (5) in the case of skewed $t$-component densities. If there were no inter-sample variation, then we could proceed to fit the same $t$- or skew $t$-mixture model simultaneously to all the $m$ samples observed. But here we have to allow for the inter-sample variation. We propose to do so by introducing random-effects terms and using them to specify how the sample-specific component distributions vary from those in the $t$- or skew $t$-mixture model representing the template.

Let $y_{ijk}$ denote the measurement on the $i$th variable for the $j$th observation in the $k$th sample ($i = 1, \ldots, p; j = 1, \ldots, n_k$; $k = 1, \ldots, m$). Then conditional on its membership of the $h$th component of the mixture model and conditional on the random-effects terms, we specify the distribution of $y_{ijk}$ as

$$y_{ijk} = a_{hik} \mu_{hi} + b_{hik} + e_{hijk}, \tag{6}$$

where $e_{hijk}$ is the error term and where $a_{hik}$ and $b_{hik}$ are random-effects terms with

$$a_{hik} \sim N(1, \xi_{1hi}^2) \quad \text{and} \quad b_{hik} \sim N(0, \xi_{2hi}^2). \tag{7}$$

Here $\boldsymbol{\mu}_{hi}$ is the $h$th component mean of the $i$th variable in the $g$-component mixture model representing the template for class $C$. The terms $e_{hijk}$, $a_{hik}$ and $b_{hik}$ are taken to be independent and this independence assumption extends over all variables and all samples. The sample-specific terms, $a_{hik}$ and $b_{hik}$, allow for scaling and translation, respectively, of the sample-component means from the component-means of the template.

We used the EM algorithm of Dempster et al.[29] to fit the $g$-component $t$- and skew $t$-mixture model with component distributions defined by (2) and (refeq5), respectively. The fitting of mixtures of $t$-distributions[27] is explained in some detail in McLachlan and Peel[30]. It is computationally convenient to use the characterization (4) to define the EM framework for the fitting of a mixture of these skew $t$-densities. However, the E-step involves the calculation of a number of conditional expectations that are unable to be expressed in closed form. Previously, Pyne et al.[7] circumvented this problem by replacing the term $\boldsymbol{\Delta}_h|\boldsymbol{U}|$ in (4) by the term $|U|\boldsymbol{\delta}_h$, where $U$ is a standard univariate normal random variable. With this simplification, the skew $t$-density (5) reduces to

$$2t_p(\boldsymbol{y}; \boldsymbol{\mu}_h, \boldsymbol{\Omega}_h, \nu_h) T_1(y^*; 0, 1, \nu_h + p), \tag{8}$$

where $\boldsymbol{y}^* = [(\nu_h + p)/\{\nu_h + d_h(\boldsymbol{y})\}\{1 - \boldsymbol{\delta}_h^T \boldsymbol{\Omega}_h^{-1} \boldsymbol{\delta}_h\}^{-1}]^{1/2} \boldsymbol{\delta}_h^T \boldsymbol{\Omega}_h^{-1}$
$(\boldsymbol{y} - \boldsymbol{\mu}_h)$. For the simplified form (8) of the skew $t$-density, the calculations on the E-step can be expressed in closed form. For the data sets analysed by Pyne et al.[7] this simplification appeared to make little difference in the fit provided by the mixture of skew $t$-distributions so modified. Hence for computational convenience we shall continue to work here with this simplified form.

In the framework of the EM algorithm for the fitting of this proposed multilevel mixture model, the complete-data are taken to be the observed samples together with the unobservable data corresponding to the variables $w$, $U$, and $U_o$ in the characterization (4) for the skew $t$-distribution, and the unknown component labels for the individual observations (cells). As observations from the same sample share common random effects terms, they will not be independently distributed. However, they will be independent conditional on the random effects, and so the complete-data log likelihood can be formed in a straightforward way for the application of the EM algorithm. Some of the conditional expectations on the E-step cannot be calculated in closed form bearing in mind that there are multiplicative random effects terms in (6) for scaling variation. We therefore proceeded by taking the multiplicative random effects to be cell-specific, i.e. using $a_{hijk}$ instead of $a_{hik}$ in (6). If the additive random effects terms were taken to be cell-specific, then they would be no longer identifiable from the error term and, so to avoid this, while they are taken to be cell-specific they are taken to be the same for each marker $i (i = 1, \ldots, m)$; that is, $b_{hik}$ in (6) is replaced by $b_{hjk}$. This means that the random-effects terms are distributed as

$$a_{hijk} \sim N(1, \xi_{1hi}^2) \quad \text{and} \quad b_{hjk} \sim N(0, \xi_{2h}^2) \quad (i = 1, \ldots, p) \tag{9}$$

for a given $h, j$, and $k \, (h = 1, \ldots, g; \, j = 1, \ldots, n_k; \, k = 1, \ldots, m)$.

The E- and M-steps, as described in Appendix A (Supplementary Information), are alternately repeated until the relative change in log likelihood is smaller than $10^{-6}$, or the number of iterations reaches 1000, whichever happens earlier. If a range for the number of populations ($g$) in the template is not specified, JCM precomputes one using the average silhouette width (implemented by pamk from the R package fpc). The selection of the optimal number of components in the mixture model is made by default on the basis of the Bayesian Information Criterion (BIC), although other criteria can be used.

**Step 3: Output files:** A list of all the input parameters ("input_parameters.txt") used for running JCM is included in the output. A tab-separated file ("batch_features.txt") contains the JCM features for every modeled sample in the batch such that each row represents a fitted parameter of the models and each column a sample. The population indices are matched across samples. The same information is also provided in GenePattern compatible format ("batch_features.gct"). A second tab-separated file ("template_features.txt") contains the features of the batch template. The template features are also provided as an R object ("template_model.ret") which can be read using the dget() function. A default list of feature-sets ("featuresets_default.gmt") is generated by pooling JCM features all panels and grouping them by types, such as population means, proportions, etc. An overlay plot of all templates across different classes ("template_overlay.pdf") and one for each class showing within-class variation ("class_overlay.pdf") are produced. Fluorescence intensity heatplots are produced for side-by-side spatial or temporal comparison of individual populations that are matched across all samples. The option of pairplots to illustrate the mixture model based clustering for each sample is also available ("sample_pairplot.pdf"). Finally, JCM outputs a zipped folder (cluster_labels.txt") containing the cluster-membership labels for each point in every sample of the batch.

**Step 4: Downstream analysis:** JCMs output formats allow easy application of visualization and other analytical routines in R, GenePattern, and other platforms. In the present analysis, for 2-class BCR signaling data, we used the Gene Set Enrichment Analysis module of GenePattern to identify the enriched meta-features (or feature-sets) across multiplexed staining panels. Although we used the featuresets_default.gmt file, in general, the user can customize feature-sets by pooling specific JCM features from selected panels and grouping them probably with the help of such feature-attributes as type, cluster # or panel #, and save the result as a .gmt file. The customization can be done easily with any text editor. (Note, both the .gct and .gmt formats are described under GenePattern file formats.) These files can then be used as input to the GSEA to identify enriched cross-panel feature-sets. Using R, BioConductor or GenePattern, one can visualize the features of interest (say, population means), or indeed all features, as heatmaps, and also select the most distinctive ones for further investigation.

## 4.1 Cell signaling datasets

Details are described in the Supplementary Information.

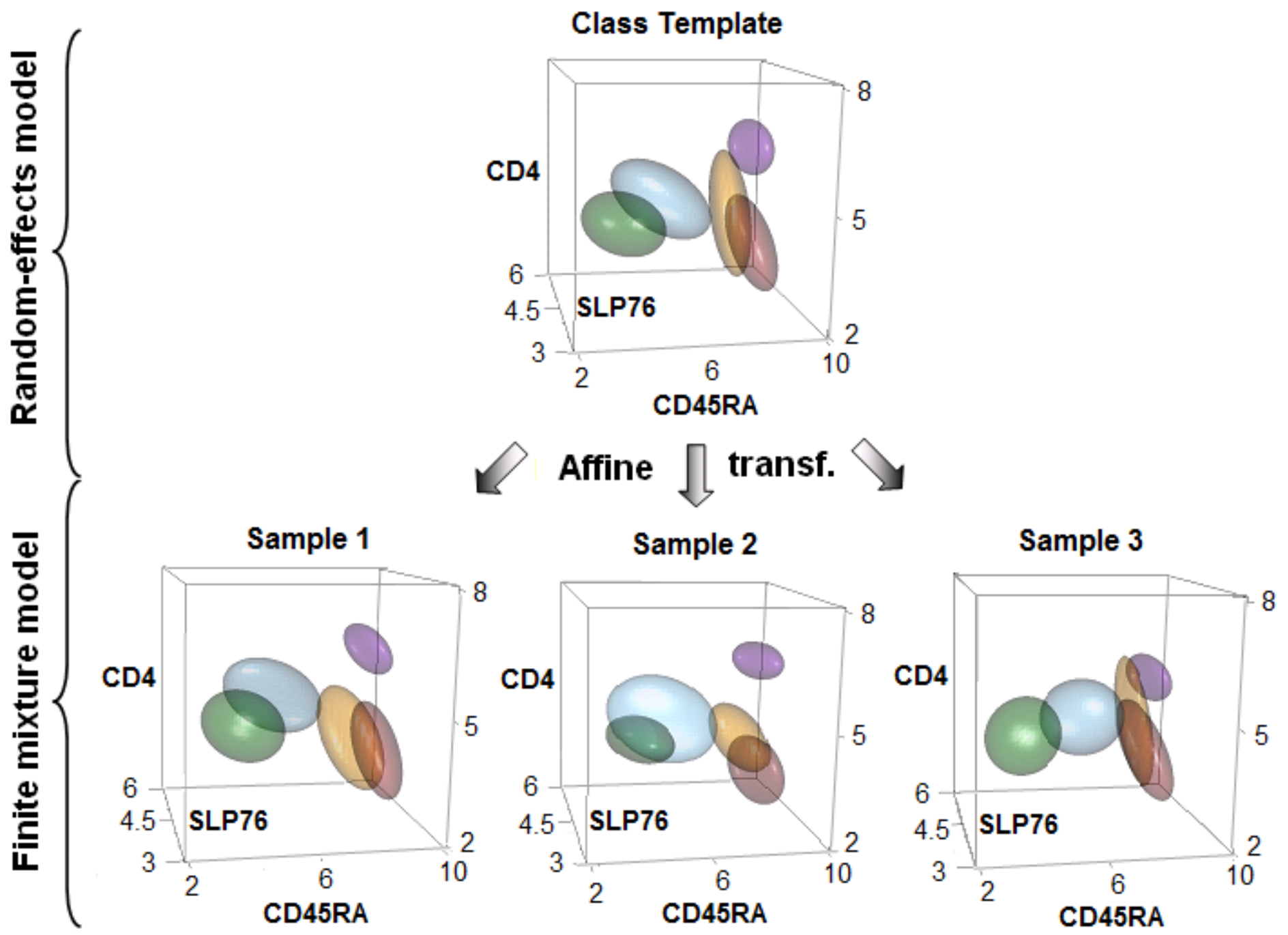

Figure 1: **JCM model and application.** The multi-level model is illustrated using the samples (bottom) and the template (top) from the 3 min class along 3 out of 4 dimensions in the data. Actual JCM-MVT parameters were used to construct these $50^{\text{th}}$ percentile multivariate $t$ density contours. The class template is computed by fitting a random effects model on all the samples, which in turn are fit with individual finite mixture models of multivariate $t$'s. Through the affine transformation parameters, each population in a sample corresponds to its counterpart in the class template, as shown by the matched colors.

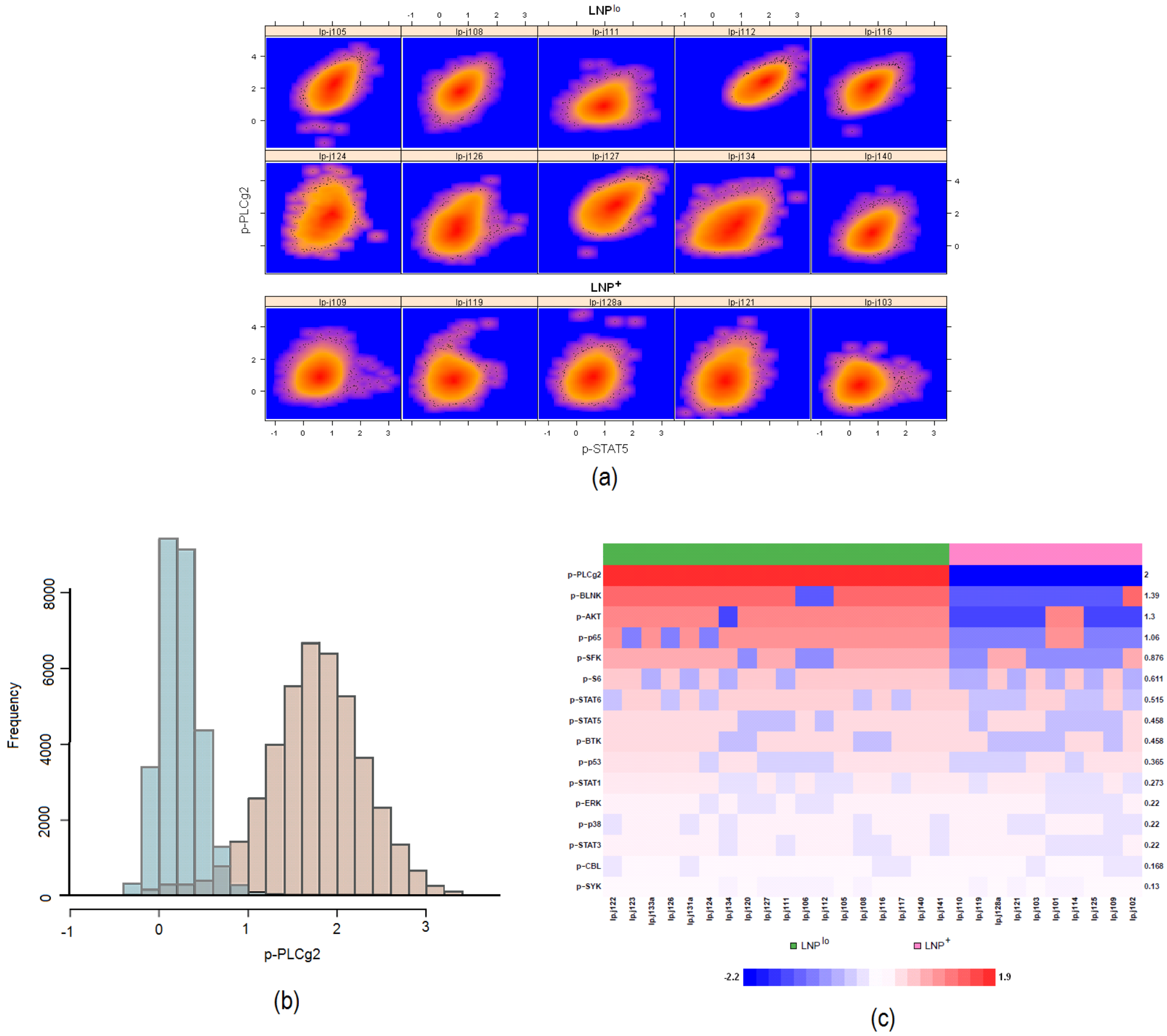

Figure 2: **Distinct spatial characteristics of phospho-marker expression in samples from two classes of patients with different outcomes.** (a) Heatplots provide insight into the distribution of phospho-proteomic expression of p-PLCg2 and p-STAT5 (panel 4) for LNP$^{lo}$ (top 2 rows) and LNP$^{+}$ (bottom row) samples. The mound (high CD20 and BCL-2) populations are shown here. In contrast to the more symmetrically distributed, well-rounded LNP$^{+}$ mounds, the skewness in the LNP$^{lo}$ mounds is clearly visible. (b) The stimulated mound (red histogram) of a LNP$^{lo}$ sample is shown in contrast with the corresponding population prior to stimulation (blue histogram). The heavy left tail of the stimulated mound probably indicates a non-LNP subpopulation with partially altered p-SFK signaling. (c) The ability of the mound skew parameters ($\boldsymbol{\delta}$) for 16 phospho-markers to distinguish samples across the LNP$^{lo}$ and LNP$^{+}$ classes (green annd pink labels respectively) is shown with a heatmap based on the corresponding posterior log-odds scores. Higher the score, darker is the corresponding entry in red/blue. Each marker name and its average posterior log-odds score over all samples are marked on the sides of the heatmap.

# Supplementary Information

## 1 Data and Experiment

### 1.1 TCR stimulation time course data

Flow-cytometry based assessment of T cell phosphorylation patterns in healthy human objects was performed as previously described [1]. While the experiments were originally conducted by the Maier et al. study [1] for the first time the time course data for all 6 classes were analyzed in high dimension by the present study. Briefly, whole blood samples from healthy Caucasian individuals were obtained with informed consent and according to the Institute Ethics Review Board protocols. Here we use the samples from three individuals that were studied over a time course of 6 time points: 0 min (pre-stimulation) and 1, 3, 5, 15 and 30 min (post-stimulation). For four-color cell surface and intercellular staining, 250,000 cells in each sample were stained using labeled antibodies CD4, CD45RA, SLP76(pY128), and ZAP70(pY292) before T cell receptor stimulation with an anti-CD3 antibody (baseline measurement; 0 min). All samples were acquired on a FACS Calibur (BD Biosciences) machine using CellQuest software within 12 hours after staining. Live cell gating was performed by Maier et al. [1] using the flowJo software. As part of data preprocessing for the present study, logicle transformation (using the flowCore package [2]) was applied to every sample; see Maier et al. [1] for further details.

### 1.2 BCR signaling data from follicular lymphoma cohorts

Follicular lymphoma (FL) tumor samples were acquired before any therapy from newly diagnosed 28 patients. All specimens were obtained with informed consent in accordance with the Declaration of Helsinki and this study was approved by Stanford University's Administrative Panels on Human Subjects in Medical Research. Tumor signaling heterogeneity was analyzed using phospho-specific flow cytometry as described in detail in the Supplemental Methods of Irish et al. [3]. Basal levels of signaling in unstimulated cells (0 minutes) were used to examine constitutive or tonic signaling. The BCR signaling response of 15 phospho-proteins was calculated as fold induction of signaling over basal at 4 minutes following stimulation by F(ab')2 against the BCR heavy chain. Lymphoma Negative Prognostic (LNP) cells, a subpopulation of lymphoma B cells found at diagnosis in some tumors, were previously shown to be continuously associated with poor overall survival [3]. The samples analyzed here constituted the Testing Set of 28 FL samples from this prior study. Three post-therapy samples from Irish et al. [3] were included to assess the effect of any inter-sample variation on JCM modeling. The 31 samples are analyzed in high dimension for the first time in the current study. LNP cells were previously quantified in these samples based on an impaired BCR signaling response in a subset of lymphoma B cells with distinct CD20 and BCL2 expression (Supplemental Methods of Irish et al. [3]). Each stimulation condition for a FL patient sample was split for staining with one of 8 staining panels which each measured a pair of phospho proteins (e.g. p-PLCg and p-STAT5) plus the lineage markers CD3, CD5, CD20, and BCL2 and cell light scatter properties. Using 8 panels allowed us to measure 16 phospho-proteins, but meant that on a given cell we only measured two phospho-proteins simultaneously. To overcome this, CD20 and BCL2 expression were used to correlate subpopulations of FL cells between staining panels [3].

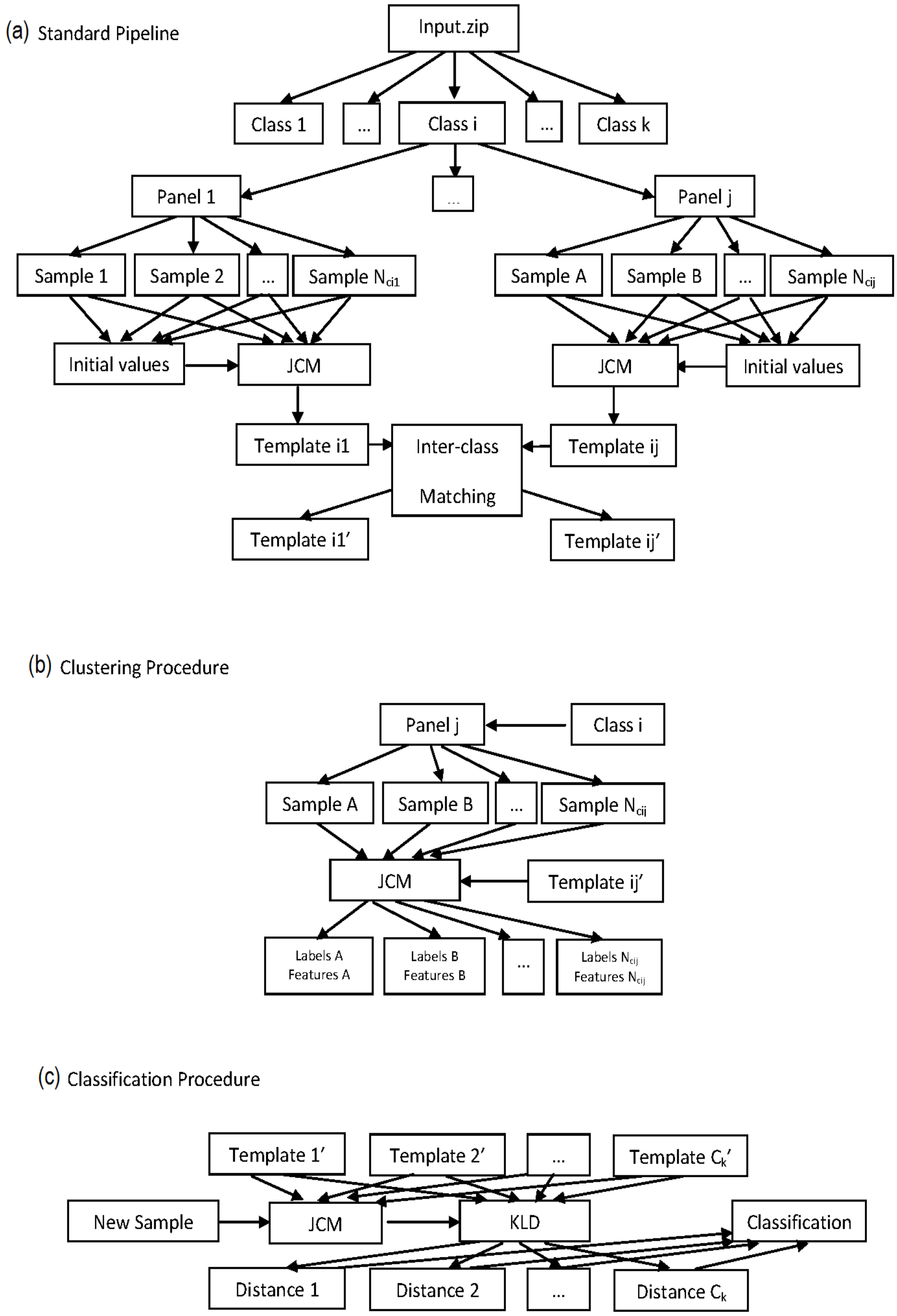

**Supplementary Figure S1**: **The Workflow of JCM.** Plot (a) shows the general workflow of JCM where the samples of each class are modeled panel-wise leading to generation of templates. Plot (b) shows specifically the clustering steps followed by JCM which results in generation of posterior probability of clustering by the mixture model for every flow event. Plot (c) shows how templates generated in the main workflow (a) could be used for classifying new samples into classes with the closest templates. This is conducted by computing the approximate high-dimensional Kullback-Leibler distance between each template and a new sample.

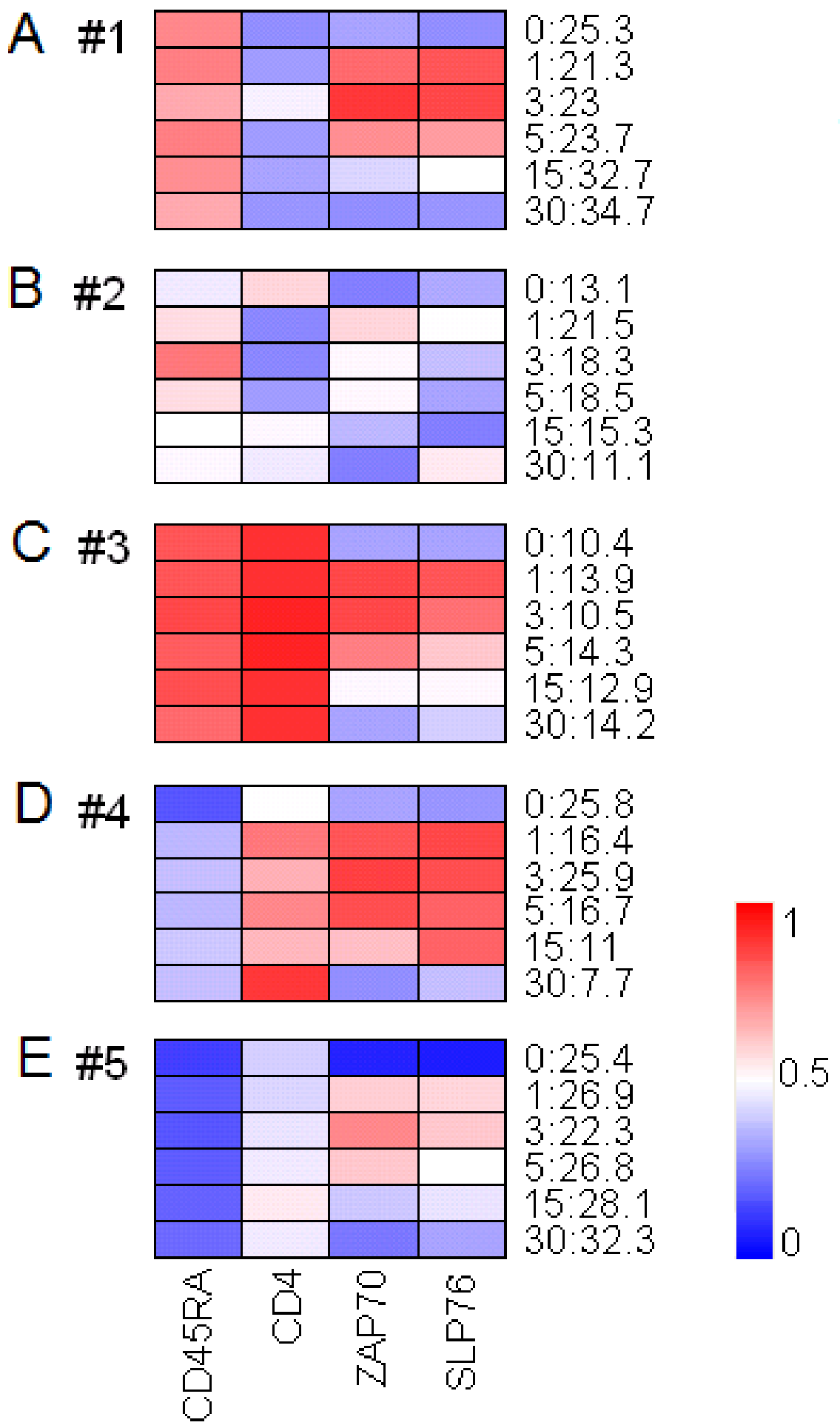

**Supplementary Figure S2**: **Spatio-temporal Characterization of Populations using JCM Class Templates.** The heatmaps depict mean intensities of 5 distinct populations (Fig. 2(a)-(e)) of the 6 class templates (rows) for all 4 markers (columns). Each class represents a time point that is indicated in the row name, followed by population size (in percentage of the total number of cells in the sample). The two rightmost columns represent phosphorylation markers. For comparison across populations, the heatmap entries are depicted in the percentile scale ranging from 0 for the lowest intensity to 1 for the highest. High intensity is shown in red, and low in blue.

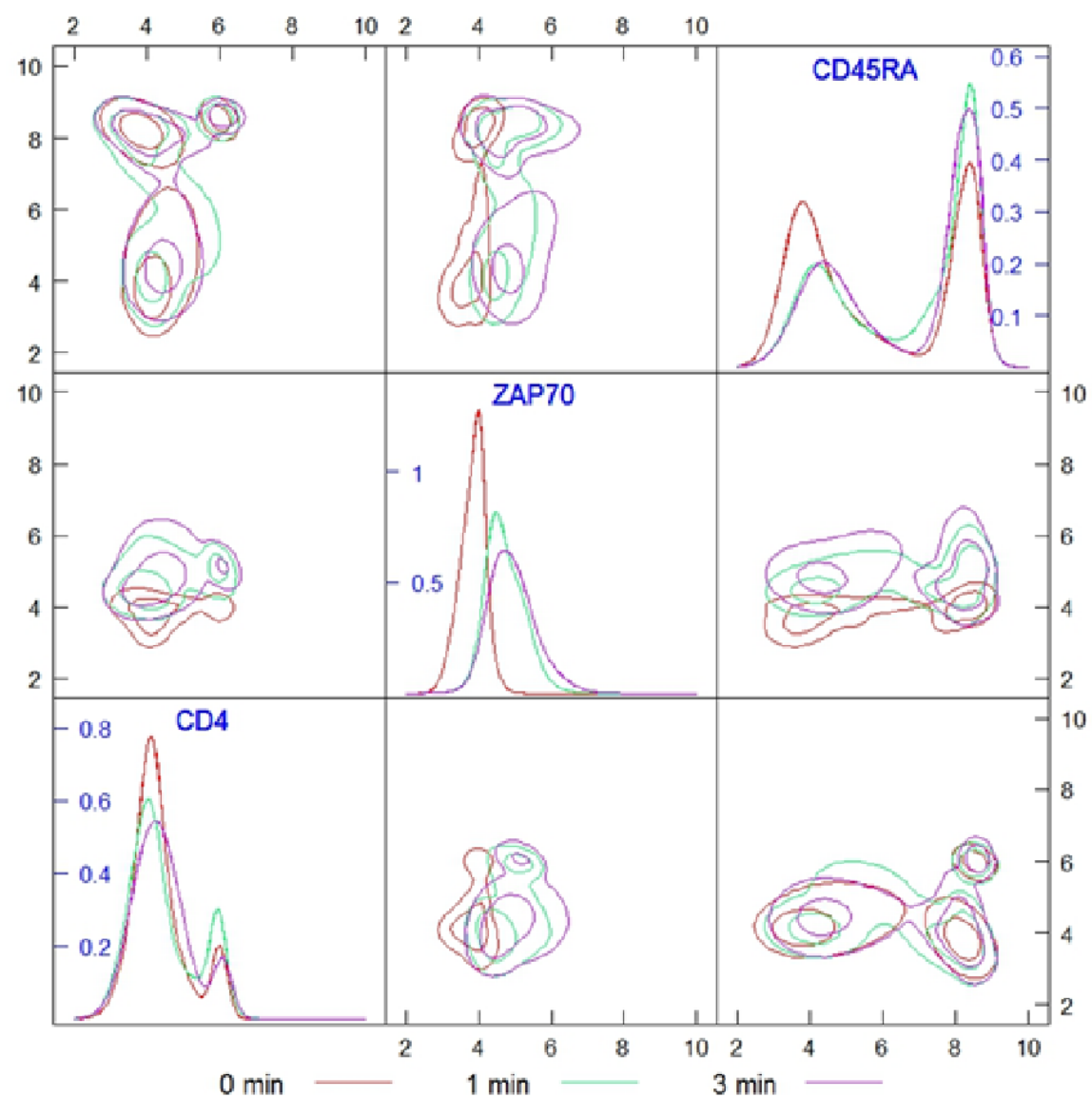

(a)

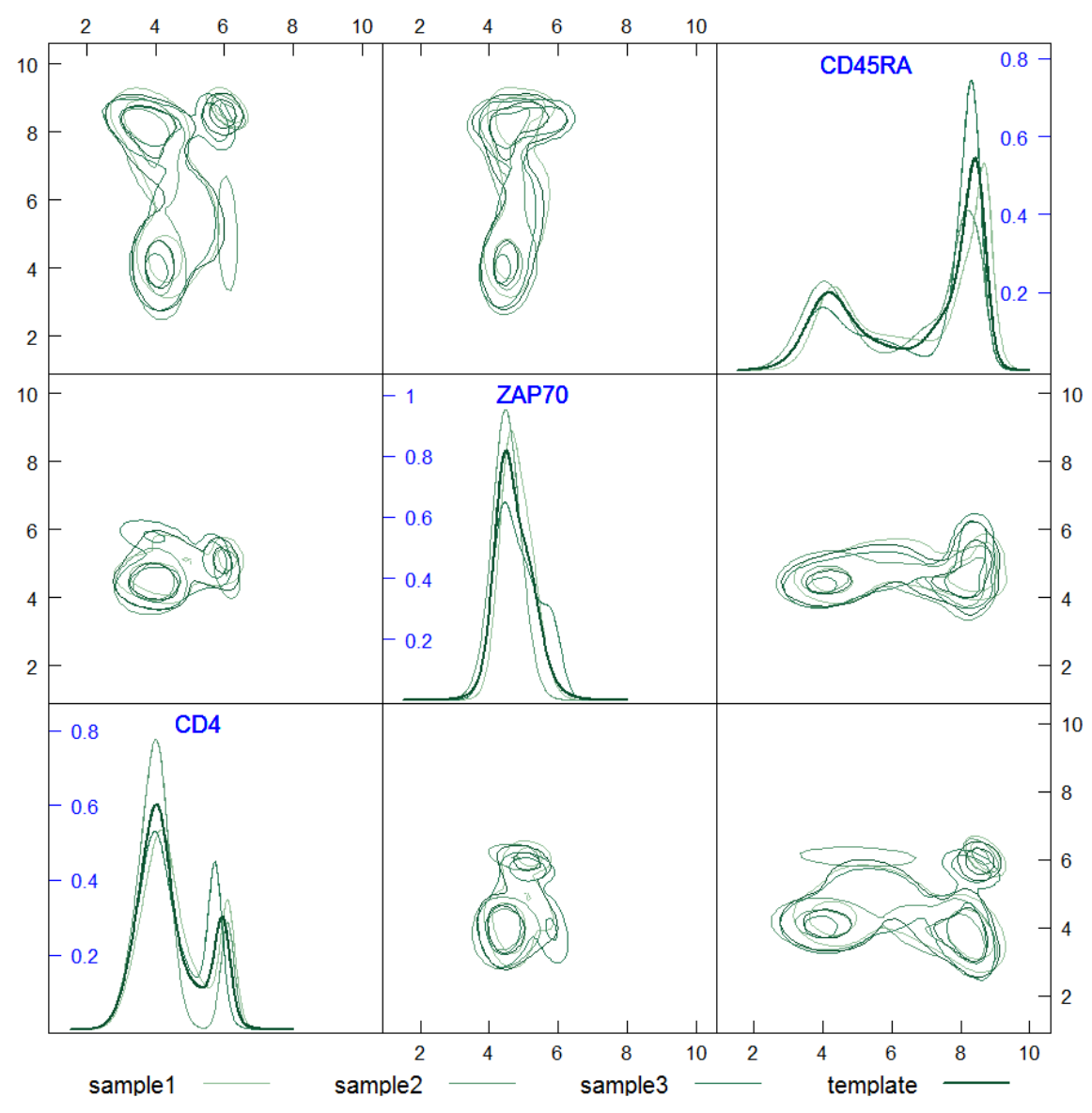

(b)

**Supplementary Figure S3**: **Overlay Plot for Capturing Variation Within a Class.** (a)An overlay plot allows spatio-temporal comparison of different class templates. The parametric templates of JCM represent each class' overall structure, and the overlay plot shows the changes therein. (b)Within-class variation among samples at $t$ = 3min (chosen arbitrarily) during TCR stimulation are overlaid for visual comparison across different markers and populations. The overall class template is shown in bold.

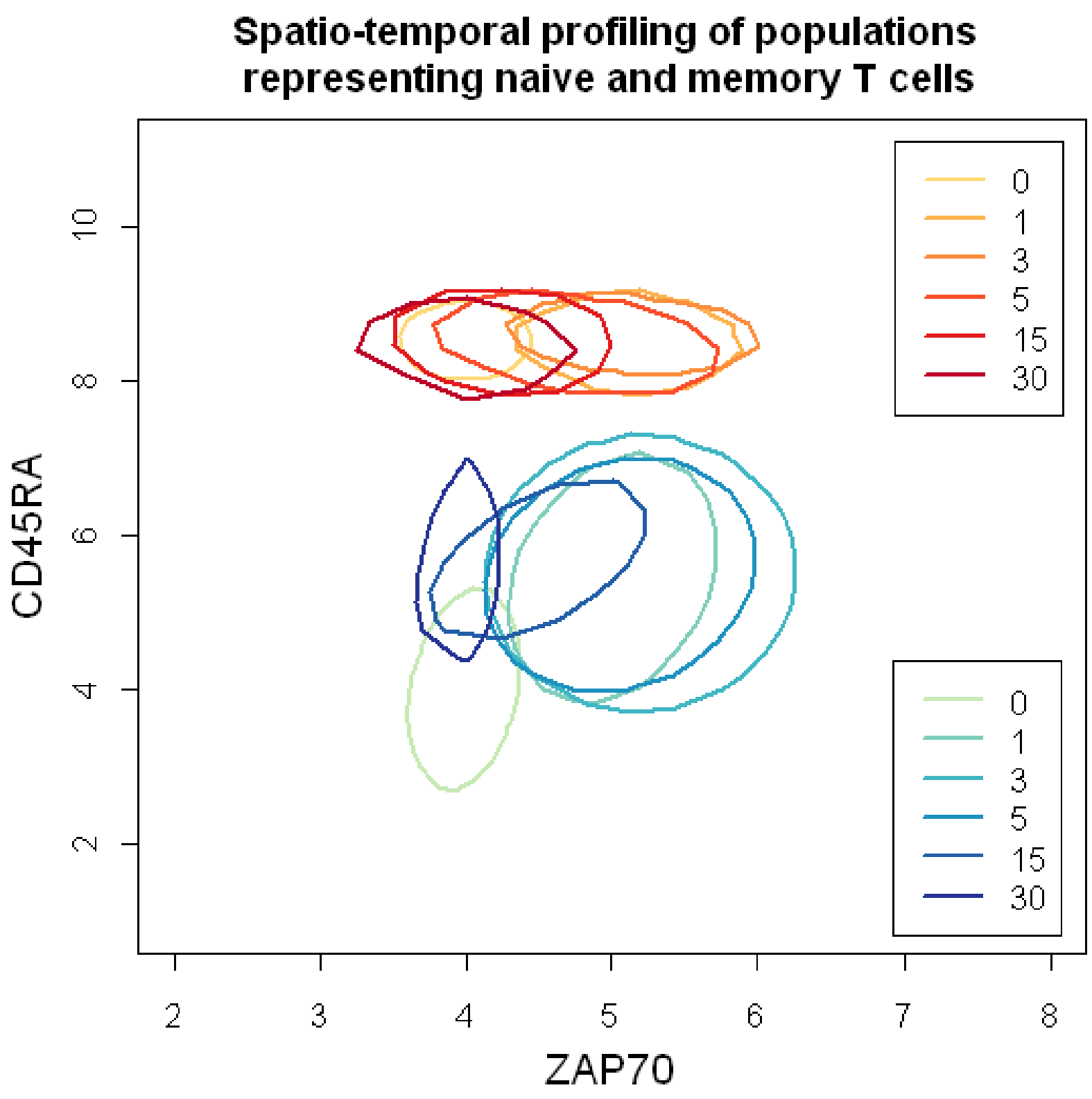

**Supplementary Figure S4**: **Spatio-temporal Profiling of Populations representing Naïve and Memory T Cells.** The sequence of templates for each of the 6 time points are shown for naïve and memory T cell subsets in reddish and bluish contours.

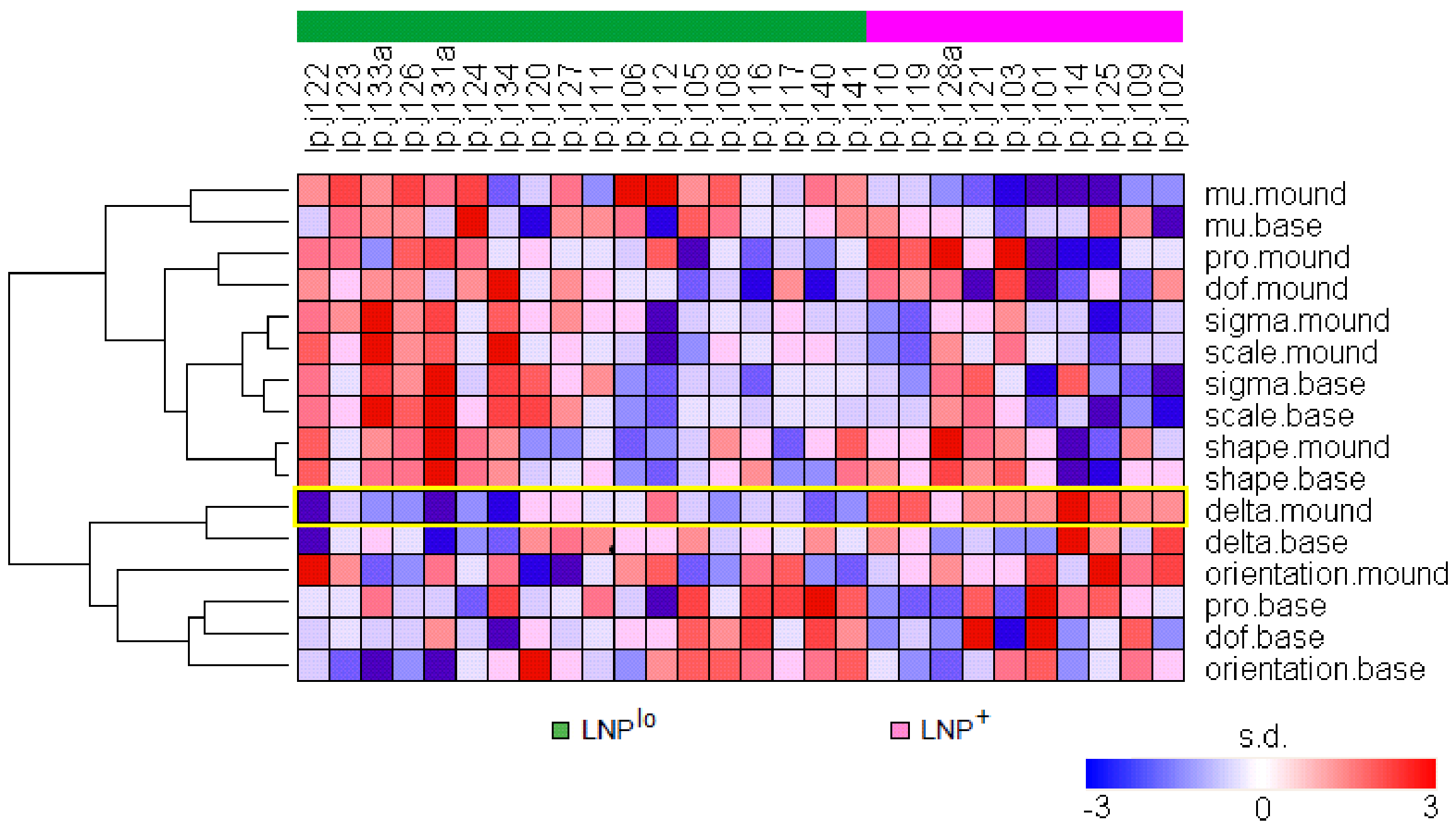

**Supplementary Figure S5**: **Enrichment of Cross-panel Meta-features.** We generalized the application of single feature enrichment analysis to metafeatures (or feature-sets) spanning all staining panels. While the individual features (such as location $\boldsymbol{\mu}$, or skewness $\boldsymbol{\delta}$) are computed with our JCM model, the feature-sets are tested for enrichment with GSEA. The high/low enrichment is shown in red/blue for each sample. The mound skew metafeature is most distinctive across $\text{LNP}^{+}$ and $\text{LNP}^{\text{lo}}$ classes.

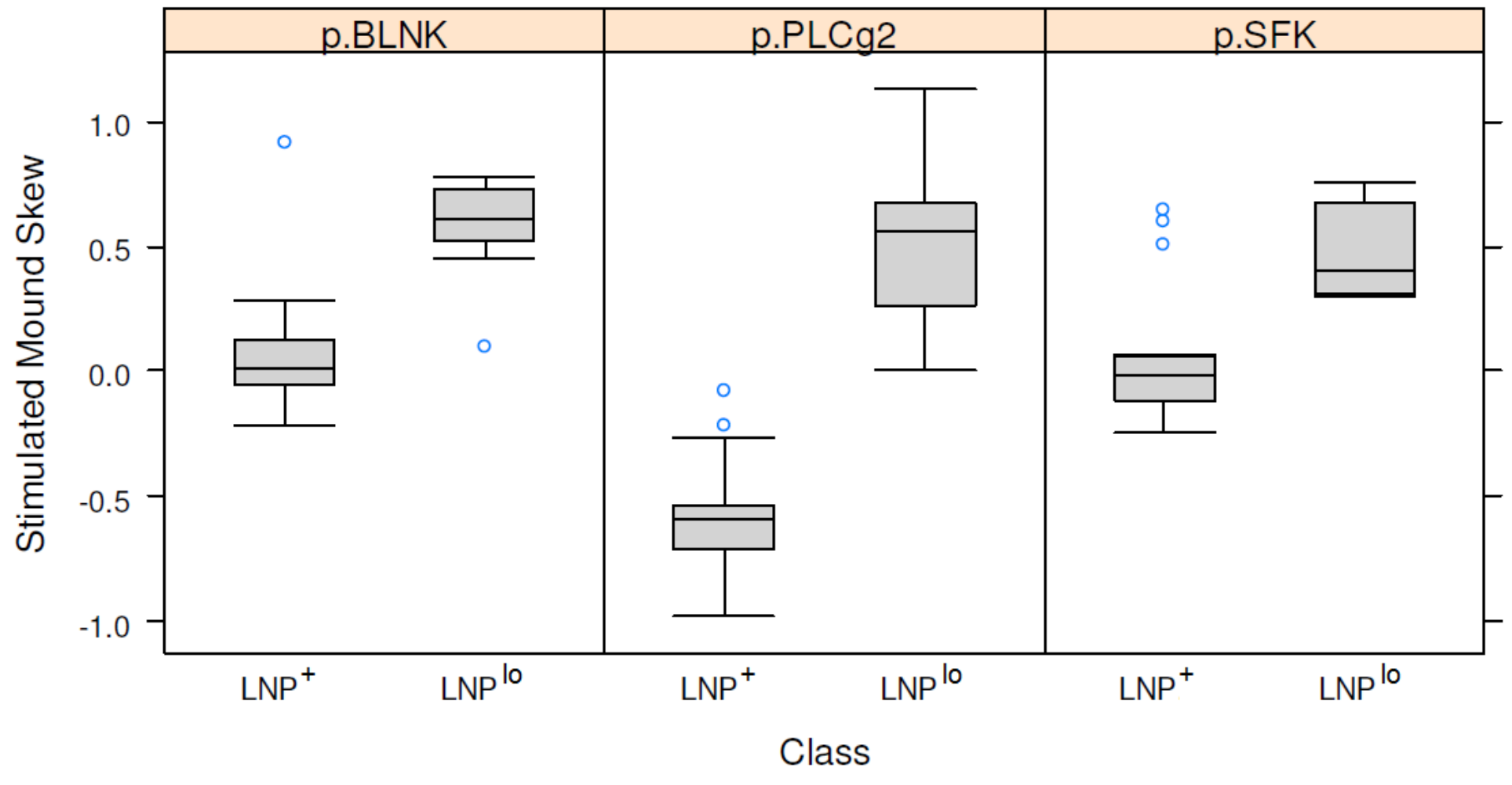

**Supplementary Figure S6**: **Differences in Mound Skewness.** The phospho-markers for which the stimulated mound (at 5 min.) is different in terms of skewness (given by parameter $\delta$ along $y$-axis) across the 2 classes of samples ($\text{LNP}^{+}$ and $\text{LNP}^{lo}$) are shown in boxplot.

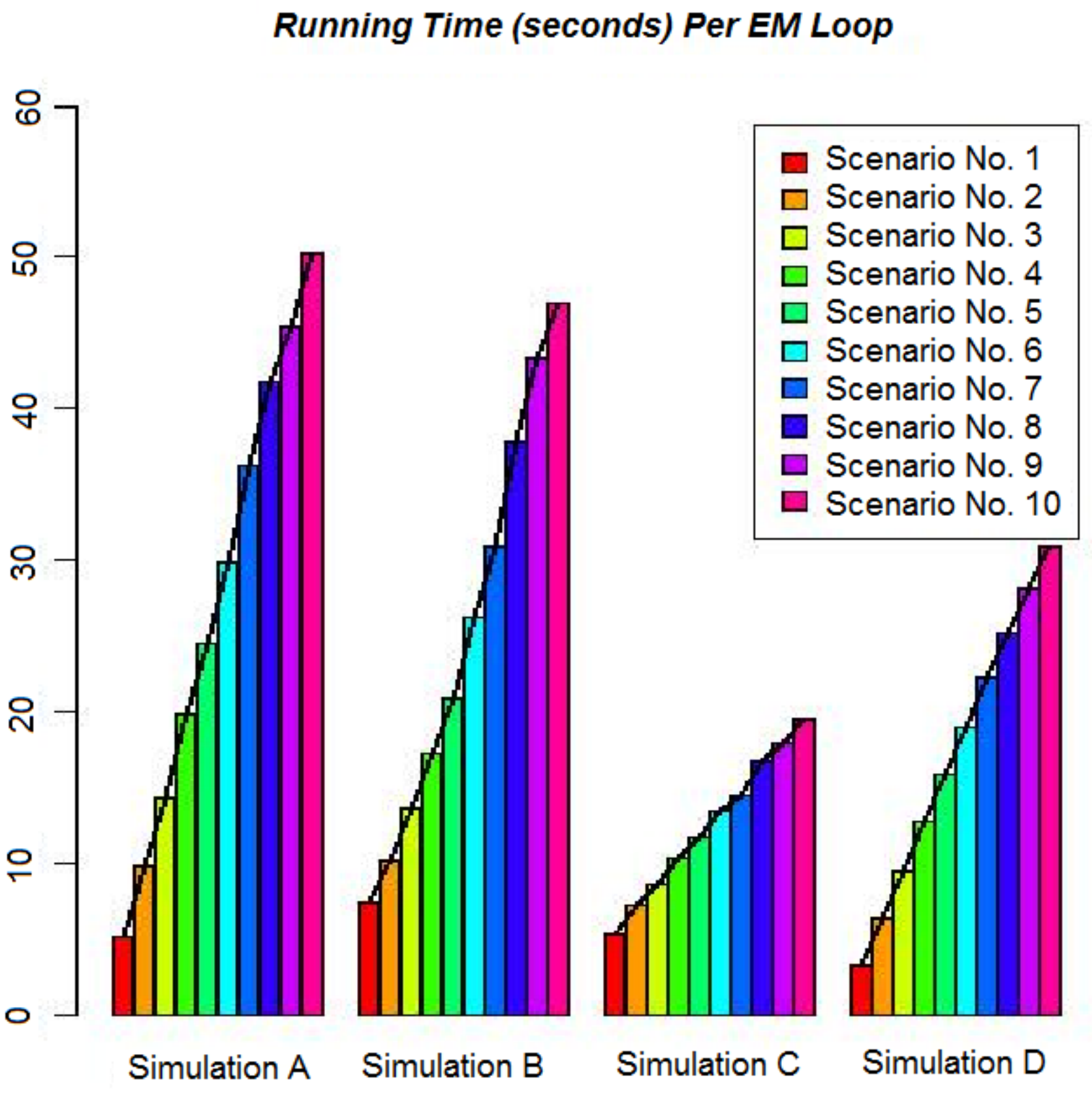

**Supplementary Figure S7**: **Running Time Analysis.** We conducted extensive stimulation studies to determine the performance of JCM for different scenarios. Simulation A was done with varying sample sizes as $n = k \times 10^3$ where $k = 10, 20, 30 \ldots 100$; Simulation B varied dimensions in data as $p = 2, 4, 6, 8, \ldots, 20$; Simulation C was run with the number of populations per sample given by $g = 1, 2, \ldots, 10$; Simulation D varied the number of samples in a cohort $m = 5, 10, \ldots, 50$. Thus we had 10 scenarios for varying every parameter. In each case, besides the parameter that was made to vary, the rest were fixed at: $n = 20,000$, $p = 4$, $g = 4$, $m = 15$. After running up to 100 iterations, the running time, which grows linearly for each of the 4 parameters, was found to be less than 1 min per EM iteration.

**Supplementary Table S1**: **Classification Error Rates of Four Methods with Benchmark Data.** Samples from different datasets from FlowCAP1 were classified using class-templates computed by four methods. These samples were pre-characterized by human experts thus allowing us to run Discriminant Analysis and compute misclassification error for each sample. Diminishing error rates are shown here in darker shades. Clearly JCM shows overall superior performance.

(A) GvHD data

| Sample | g | FLAME | pooled | JCM | flowClust |
|---|---|---|---|---|---|
| Sa001 | 4 | 0.125 | 0.111 | 0.118 | 0.170 |
| Sa002 | 5 | 0.342 | 0.273 | 0.264 | 0.256 |
| Sa003 | 3 | 0.392 | 0.001 | 0.002 | 0.054 |

(B) CFSE data

| Sample | g | FLAME | pooled | JCM | flowClust |
|---|---|---|---|---|---|
| Sa001 | 4 | 0.275 | 0.266 | 0.260 | 0.401 |
| Sa002 | 4 | 0.231 | 0.241 | 0.237 | 0.179 |
| Sa003 | 4 | 0.226 | 0.206 | 0.203 | 0.215 |

(C) DLBCL data

| Sample | g | FLAME | pooled | JCM | flowClust |
|---|---|---|---|---|---|
| Sa001 | 2 | 0.103 | 0.015 | 0.015 | 0.106 |
| Sa002 | 3 | 0.112 | 0.123 | 0.123 | 0.068 |
| Sa003 | 4 | 0.074 | 0.078 | 0.077 | 0.076 |

(D) StemCell data

| Sample | g | FLAME | pooled | JCM | flowClust |
|---|---|---|---|---|---|
| Sa001 | 4 | 0.000 | 0.000 | 0.000 | 0.000 |
| Sa002 | 5 | 0.006 | 0.002 | 0.002 | 0.145 |
| Sa003 | 3 | 0.375 | 0.004 | 0.004 | 0.018 |

# Appendix A The JCM-MVT Model

We describe in this appendix the EM algorithm for the model described in Section 4 with the family of multivariate $t$-densities as the component distributions. The description is presented for a given patient $k\,(k = 1, \ldots, m)$. For convenience of notation, we shall suppress the subscript $k$ from the random-effects terms $a_{hijk}$ and $b_{hjk}$ as specified by (9). We let $\boldsymbol{\Psi}_h$ be the diagonal matrix with $i$th diagonal element equal to $\xi_{1hi}^2\,(i = 1, \ldots, p)$ for $h = 1, \ldots, g$.

## A.1 E-step

The EM algorithm is applied iteratively with the E- and M-steps alternate repeatedly. On the $(r+1)$th iteration, the E-step requires the calculation of the complete-data log likelihood given the observed data, using the current estimate of the parameter vector. This requires the following conditional expectations to be evaluated,

$$\begin{aligned}
z_{hj}^{(r)} &= E_{\boldsymbol{\Psi}^{(r)}}(z_{hj} = 1 \mid \boldsymbol{y}_j), &(1)\\
u_{hj}^{(r)} &= E_{\boldsymbol{\Psi}^{(r)}}(w_{hj} \mid \boldsymbol{y}_j, z_{hj} = 1), &(2)\\
\boldsymbol{S}_{1,hj}^{(r)} &= E_{\boldsymbol{\Psi}^{(r)}}(w_{hj}(\boldsymbol{a}_{hj} - \boldsymbol{1}_p)(\boldsymbol{a}_{hj} - \boldsymbol{1}_p)^T \mid \boldsymbol{y}_j, z_{hj} = 1), &(3)\\
S_{2,hj}^{(r)} &= E_{\boldsymbol{\Psi}^{(r)}}(w_{hj} b_{hj}^2 \mid, \boldsymbol{y}_j, z_{hj} = 1), &(4)\\
\boldsymbol{S}_{3,hj}^{(r)} &= E_{\boldsymbol{\Psi}^{(r)}}(w_{hj}(\boldsymbol{y}_j - \boldsymbol{\zeta}_h^{(r)}\boldsymbol{a}_{hj} - \boldsymbol{1}_p b_{hj})(\boldsymbol{y}_j - \boldsymbol{\zeta}_h^{(r)}\boldsymbol{a}_{hj} - \boldsymbol{1}_p b_{hj})^T \mid \boldsymbol{y}_j, z_{hj} = 1), &(5)
\end{aligned}$$

where

$$\boldsymbol{\zeta}_h^{(r)} = \mathrm{DIAG}(\boldsymbol{\mu}_h^{(r)}).$$

We use $\mathrm{DIAG}(\boldsymbol{a})$ to denote the $p \times p$ diagonal matrix which has diagonal elements given by the $p$-dimensional vector $\boldsymbol{a}$; we use $\mathrm{diag}(\boldsymbol{A})$ to denote vector given by the diagonal elements of the matrix $\boldsymbol{A}$. For brevity of notation, we suppress the fact that $z_{hj} = 1$ in our equations in the sequel.

It is easy to show that the first two conditional expectations are given by

$$z_{hj}^{(r)} = \frac{\pi_h^{(r)} t_p(\boldsymbol{y}_j; \boldsymbol{\mu}_h^{(r)}, \boldsymbol{\Omega}_h^{(r)}, \nu_h^{(r)})}{\sum_{h=1}^{g} \pi_h^{(r)} t_p(\boldsymbol{y}_j; \boldsymbol{\mu}_h^{(r)}, \boldsymbol{\Omega}_h^{(r)}, \nu_h^{(r)})}, \tag{6}$$

and

$$u_{hj}^{(r)} = \frac{\nu_h^{(r)} + p}{\nu_h^{(r)} + d_h^{(r)}(\boldsymbol{y}_j)}, \tag{7}$$

where

$$\boldsymbol{\Omega}_h^{(r)} = \boldsymbol{\zeta}_h^{(r)} \boldsymbol{\Psi}_h^{(r)} \boldsymbol{\zeta}_h^{(r)^T} + \xi_{2h}^{(r)^2} \boldsymbol{1}_p \boldsymbol{1}_p^T + \boldsymbol{\Sigma}_h^{(r)}$$

and

$$d_h^{(r)}(\boldsymbol{y}_j) = (\boldsymbol{y}_j - \boldsymbol{\mu}_h^{(r)})^T \boldsymbol{\Omega}_h^{(r)^{-1}} (\boldsymbol{y}_j - \boldsymbol{\mu}_h^{(r)}).$$

To obtain the remaining four conditional expectations, we use the following distributional result,

$$\left[\begin{array}{c} \boldsymbol{a}_{hj} \\ b_{hj} \\ \boldsymbol{y}_j \end{array}\right] \mid w_{hj} \sim \mathrm{N}_{2p+1}\left(\left[\begin{array}{c} \mathbf{1}_p \\ 0 \\ \boldsymbol{\mu}_h \end{array}\right], \left[\begin{array}{ccc} \boldsymbol{\Psi}_h & \mathbf{0}_p & \boldsymbol{\Psi}_h \boldsymbol{\zeta}_h^T \\ \mathbf{0}_p^T & \xi_{2h}^2 & \xi_{2h}^2 \mathbf{1}_p^T \\ \boldsymbol{\zeta}_h \boldsymbol{\Psi}_h & \xi_{2h}^2 \mathbf{1}_p & \boldsymbol{\Omega}_h \end{array}\right] \frac{1}{w_{hj}}\right). \tag{8}$$

Conditional on $\boldsymbol{y}_j$ and $w_{hj}$, it follows that $[\boldsymbol{a}_{hj} b_{hj}]^T$ has covariance matrix given by

$$\begin{aligned} \boldsymbol{V}_{1,hj}^{(r)} &= \mathrm{cov}_{\boldsymbol{\Psi}^{(r)}}\left(\left[\begin{array}{c} \boldsymbol{a}_{hj} \\ b_{hj} \end{array}\right] \mid \boldsymbol{y}_j, w_{hj}\right) \\ &= \frac{1}{w_{hj}}\left[\begin{array}{cc} \boldsymbol{\Psi}_h^{(r)} & 0 \\ 0 & \xi_h^{(r)^2} \end{array}\right] - \frac{1}{w_{hj}}\left[\begin{array}{c} \boldsymbol{\Psi}_h^{(r)} \boldsymbol{\zeta}_h^{(r)^T} \\ \xi_h^{(r)^2} \mathbf{1}_p^T \end{array}\right] \boldsymbol{\Omega}_h^{(r)} \left[\boldsymbol{\zeta}_h^{(r)} \boldsymbol{\Psi}_h^{(r)} \quad \xi_h^{(r)^2} \mathbf{1}_p\right]. \end{aligned} \tag{9}$$

From (9), we have that

$$\begin{aligned} E_{\boldsymbol{\Psi}^{(r)}}\left(\boldsymbol{a}_{hj} - \mathbf{1}_p \mid \boldsymbol{y}_j, w_{hj}\right) &= \boldsymbol{\Psi}_h^{(r)} \boldsymbol{\zeta}_h^{(r)^T} \boldsymbol{\Omega}_h^{(r)^{-1}} \left(\boldsymbol{y}_j - \boldsymbol{\mu}_h^{(r)}\right), & (10) \\ E_{\boldsymbol{\Psi}^{(r)}}(b_{hj} \mid \boldsymbol{y}_j, w_{hj}) &= \xi_{2h}^{(r)^2} \mathbf{1}_p^T \boldsymbol{\Omega}_h^{(r)^{-1}} \left(\boldsymbol{y}_j - \boldsymbol{\mu}_h^{(r)}\right), & (11) \\ \mathrm{cov}_{\boldsymbol{\Psi}^{(r)}}\left(\boldsymbol{a}_{hj} - \mathbf{1}_p \mid \boldsymbol{y}_j, w_{hj}\right) &= \frac{1}{w_{hj}}\left(\boldsymbol{\Psi}_h^{(r)} - \boldsymbol{\Psi}_h^{(r)} \boldsymbol{\zeta}_h^{(r)} \boldsymbol{\Omega}_h^{(r)^{-1}} \boldsymbol{\zeta}_h^{(r)} \boldsymbol{\Psi}_h^{(r)}\right), & (12) \\ \mathrm{cov}_{\boldsymbol{\Psi}^{(r)}}(b_{hj} \mid \boldsymbol{y}_j, w_{hj}) &= \frac{1}{w_{hj}}\left(\xi_{2h}^{(r)^2} - \xi_{2h}^{(r)^2} \mathbf{1}_p^T \boldsymbol{\Omega}_h^{(r)^{-1}} \mathbf{1}_p\right), & (13) \\ \mathrm{cov}_{\boldsymbol{\Psi}^{(r)}}\left(\boldsymbol{a}_{hj} b_{hj} \mid \boldsymbol{y}_j, w_{hj}\right) &= -\frac{1}{w_{hj}}\left(\xi_{2h}^{(r)^2} \boldsymbol{\Psi}_h^{(r)} \boldsymbol{\zeta}_h^{(r)^T} \boldsymbol{\Omega}_h^{(r)^{-1}} \mathbf{1}_p\right). & (14) \end{aligned}$$

By noting that $E(\boldsymbol{X}\boldsymbol{X}^T) = \mathrm{cov}(\boldsymbol{X}) + E(\boldsymbol{X})E(\boldsymbol{X})^T$, the conditional expectations (3), (4), and (5) are given by

$$\boldsymbol{S}_{1,hj}^{(r)} = \boldsymbol{\Psi}_h^{(r)} - \boldsymbol{\Psi}_{ah}^{(r)} \boldsymbol{\zeta}_h^{(r)^T} \boldsymbol{V}_{2,hj}^{(r)} \boldsymbol{\zeta}_h^{(r)} \boldsymbol{\Psi}_h^{(r)}, \tag{15}$$

$$\boldsymbol{S}_{2,hj}^{(r)} = \xi_{2h}^{(r)^2} - \xi_{2h}^{(r)^2} \mathbf{1}_p^T \boldsymbol{V}_{2,hj}^{(r)} \mathbf{1}_p, \tag{16}$$

where

$$\boldsymbol{V}_{2,hj}^{(r)} = \left(\boldsymbol{\Omega}_h^{(r)^{-1}} - u_{hj}^{(r)} \boldsymbol{\Omega}_h^{(r)^{-1}} \left(\boldsymbol{y}_j - \boldsymbol{\mu}_h^{(r)}\right) \left(\boldsymbol{y}_j - \boldsymbol{\mu}_h^{(r)}\right)^T \boldsymbol{\Omega}_h^{(r)^{-1}}\right) \tag{17}$$

and

$$\begin{aligned} \boldsymbol{S}_{3,hj}^{(r)} &= E_{\boldsymbol{\Psi}^{(r)}}\left(w_{hj} \left[\boldsymbol{\zeta}_h^{(r)} \quad \mathbf{1}_p\right] \boldsymbol{V}_{1,hj}^{(r)} \left[\begin{array}{c} \boldsymbol{\zeta}_h^{(r)^T} \\ \mathbf{1}_p^T \end{array}\right] \mid \boldsymbol{y}_j\right) + u_{hj}^{(r)} \boldsymbol{V}_{3,hj}^{(r)} \boldsymbol{V}_{3,hj}^{(r)^T} \\ &= \boldsymbol{V}_{4,hj}^{(r)} - \boldsymbol{V}_{4,hj}^{(r)} \boldsymbol{\Omega}_h^{(r)^{-1}} \boldsymbol{V}_{4,hj}^{(r)} + u_{hj}^{(r)} \boldsymbol{V}_{3,hj}^{(r)} \boldsymbol{V}_{3,hj}^{(r)^T}, \end{aligned} \tag{18}$$

where

$$\boldsymbol{V}_{3,hj}^{(r)} = \boldsymbol{y}_j - \boldsymbol{\mu}_h^{(r)} - \left(\boldsymbol{\zeta}_h^{(r)} \boldsymbol{\Psi}_h^{(r)} + \xi_{2h}^{(r)^2} \mathbf{1}_p\right) \boldsymbol{\Omega}_h^{(r)^{-1}} \left(\boldsymbol{y}_j - \boldsymbol{\mu}_h^{(r)}\right) \tag{19}$$

and

$$\boldsymbol{V}_{4,hj}^{(r)} = \boldsymbol{\zeta}_h^{(r)} \boldsymbol{\Psi}_h \boldsymbol{\zeta}_h^{(r)} + \xi_{2h}^{(r)^2} \mathbf{1}_p \mathbf{1}_p^T. \tag{20}$$

In order to calculate $\boldsymbol{\mu}_h^{(r+1)}$, we need to calculate two additional quantities,

$$\boldsymbol{S}_{4,hj}^{(r)} = E_{\boldsymbol{\Psi}^{(r)}} \left( w_{hj} E_{\boldsymbol{\Psi}^{(r)}} \left( \boldsymbol{A}_{hj} \boldsymbol{\Sigma}_h^{(r)^{-1}} \boldsymbol{A}_{hj} \mid \boldsymbol{y}_j, w_{hj} \right) \mid \boldsymbol{y}_j \right) \tag{21}$$

and

$$\boldsymbol{S}_{5,hj}^{(r)} = E_{\boldsymbol{\Psi}^{(r)}} \left( w_{hj} E_{\boldsymbol{\Psi}^{(r)}} \left( \boldsymbol{A}_{hj} \boldsymbol{\Sigma}_h^{(r)^{-1}} \left( \boldsymbol{y}_j - \mathbf{1}_p b_{hj} \right) \mid \boldsymbol{y}_j, w_{hj} \right) \mid \boldsymbol{y}_j \right), \tag{22}$$

where $\boldsymbol{A}_{hj}$ denotes that diagonal matrix with $\boldsymbol{a}_{hj}$ as its diagonal elements.

Note that

$$E_{\boldsymbol{\Psi}}^{(r)} \left( \boldsymbol{A}_{hj} \boldsymbol{\Sigma}_h^{(r)^{-1}} \boldsymbol{A}_{hj} \mid \boldsymbol{y}_j, w_{hj} \right) = E_{\boldsymbol{\Psi}}^{(r)} \left( \boldsymbol{a}_{hj} \boldsymbol{a}_{hj}^T \mid \boldsymbol{y}_j, w_{hj} \right) \odot \boldsymbol{\Sigma}_h^{(r)^{-1}}, \tag{23}$$

where $\odot$ denotes the elementwise matrix multiplication. Observe that

$$\begin{aligned} E_{\boldsymbol{\Psi}^{(r)}} \left( \boldsymbol{a}_{hj} \boldsymbol{a}_{hj}^T \mid \boldsymbol{y}_j, w_{hj} \right) &= \frac{1}{w_{hj}} \left( \boldsymbol{\Psi}_h^{(r)} - \boldsymbol{\Psi}_h^{(r)} \boldsymbol{\zeta}_h^{(r)^T} \boldsymbol{\Omega}_h^{(r)^{-1}} \boldsymbol{\zeta}_h^{(r)} \boldsymbol{\Psi}_h^{(r)} \right) \\ &\quad + \left[ \mathbf{1}_p + \boldsymbol{\Psi}_h^{(r)} \boldsymbol{\zeta}_h^{(r)^T} \boldsymbol{\Omega}_h^{(r)^{-1}} \left( \boldsymbol{y}_j - \boldsymbol{\mu}_h^{(r)} \right) \right] \\ &\quad \times \left[ \mathbf{1}_p^T + \left( \boldsymbol{y}_j - \boldsymbol{\mu}_h^{(r)} \right)^T \boldsymbol{\Omega}_h^{(r)^{-1}} \boldsymbol{\zeta}_h^{(r)} \boldsymbol{\Psi}_h^{(r)} \right]. \end{aligned} \tag{24}$$

It follows that

$$\boldsymbol{S}_{4,hj}^{(r)} = \boldsymbol{V}_{5,hj}^{(r)} \odot \boldsymbol{\Sigma}_h^{(r)^{-1}}, \tag{25}$$

where

$$\begin{aligned} \boldsymbol{V}_{5,hj}^{(r)} &= \left( \boldsymbol{\Psi}_h^{(r)} - \boldsymbol{\Psi}_h^{(r)} \boldsymbol{\zeta}_h^{(r)} \boldsymbol{\Omega}_h^{(r)^{-1}} \boldsymbol{\zeta}_h^{(r)} \boldsymbol{\Psi}_h^{(r)} \right) \\ &\quad + u_{hj}^{(r)} \left[ \mathbf{1}_p + \boldsymbol{\Psi}_h^{(r)} \boldsymbol{\zeta}_h^{(r)} \boldsymbol{\Omega}_h^{(r)^{-1}} \left( \boldsymbol{y}_j - \boldsymbol{\mu}_h^{(r)} \right) \right] \\ &\quad \times \left[ \mathbf{1}_p + \boldsymbol{\Psi}_h^{(r)} \boldsymbol{\zeta}_h^{(r)} \boldsymbol{\Omega}_h^{(r)^{-1}} \left( \boldsymbol{y}_j - \boldsymbol{\mu}_h^{(r)} \right) \right]^T. \end{aligned} \tag{26}$$

To calculate $\boldsymbol{S}_{5,hj}^{(r)}$, we note that

$$\begin{aligned} &= \boldsymbol{E}_{\boldsymbol{\Psi}^{(r)}} \left( \boldsymbol{A}_{hj} \boldsymbol{\Sigma}_h^{(r)^{-1}} \left( \boldsymbol{y}_j - \mathbf{1}_p b_{hj} \right) \mid \boldsymbol{y}_j, w_{hj} \right) \\ &= E_{\boldsymbol{\Psi}}^{(r)} \left( \boldsymbol{A}_{hj} \mid \boldsymbol{y}_j, w_{hj} \right) \boldsymbol{\Sigma}_h^{(r)^{-1}} \boldsymbol{y}_j - E_{\boldsymbol{\Psi}^{(r)}} \left( \boldsymbol{A}_{hj} b_{hj} \mid \boldsymbol{y}_j, w_{hj} \right) \boldsymbol{\Sigma}_h^{(r)^{-1}} \mathbf{1}_p. \end{aligned}$$

Then

$$\begin{aligned} \boldsymbol{S}_{5,hj}^{(r)} &= u_{hj}^{(r)} \boldsymbol{V}_{6,hj}^{(r)} \boldsymbol{\Sigma}_h^{(r)^{-1}} \boldsymbol{y}_j + \text{ DIAG} \left( \xi_{2h}^{(r)^2} \boldsymbol{\Psi}_h^{(r)} \boldsymbol{\zeta}_h^{(r)} \boldsymbol{\Omega}_h^{(r)^{-1}} \mathbf{1}_p \right) \boldsymbol{\Sigma}_h^{(r)^{-1}} \mathbf{1}_p, \\ &\quad - u_{hj}^{(r)} \text{DIAG} \left[ \xi_{2h}^{(r)^2} \boldsymbol{V}_{6,hj}^{(r)} \mathbf{1}_p^T \boldsymbol{\Omega}_h^{(r)^{-1}} \left( \boldsymbol{y}_j - \boldsymbol{\mu}_h^{(r)} \right) \right] \boldsymbol{\Sigma}_h^{(r)^{-1}} \mathbf{1}_p, \end{aligned} \tag{27}$$

where

$$\boldsymbol{V}_{6,hj}^{(r)} = \text{DIAG} \left( \mathbf{1}_p + \boldsymbol{\Psi}_h^{(r)} \boldsymbol{\zeta}_h^{(r)} \boldsymbol{\Omega}_h^{(r)^{-1}} \left( \boldsymbol{y}_j - \boldsymbol{\mu}_h^{(r)} \right) \right). \tag{28}$$

## A.2 M-step

The estimates of the parameters are updated on the M-step by maximizing the $Q$-function over the parameter space. The $Q$-function is equal to the conditional expectation of the complete-data log likelihood given the observed data, using the current fit for the vector of unknown parameters. It follows that

$$\begin{aligned}
\pi_h^{(r+1)} &= \frac{1}{n}\sum_{j=1}^{n} z_{hj}^{(r)}, \\
\mu_h^{(r+1)} &= \left(\sum_{j=1}^{n} z_{hj}^{(r)} \boldsymbol{S}_{4,hj}^{(r)}\right)^{-1} \sum_{j=1}^{n} z_{hj}^{(r)} \boldsymbol{S}_{5,hj}^{(r)}, \\
\boldsymbol{\Sigma}_h^{(r+1)} &= \frac{\sum_{j=1}^{n} z_{hj}^{(r)} \boldsymbol{S}_{3,hj}^{(r)}}{\sum_{j=1}^{n} z_{hj}^{(r)}}, \\
\boldsymbol{\Psi}_h^{(r+1)} &= \frac{\sum_{j=1}^{n} z_{hj}^{(r)} \boldsymbol{S}_{1,hj}^{(r)}}{\sum_{j=1}^{n} z_{hj}^{(r)}}, \\
\xi_{2h}^{(k+1)^2} &= \frac{\sum_{j=1}^{n} z_{hj}^{(r)} S_{2,hj}^{(r)}}{\sum_{j=1}^{n} z_{hj}^{(r)}}.
\end{aligned} \tag{29}$$

The update of the degrees of freedom $\nu_h^{(r+1)}$ is given implicitly as a solution of the equation

$$\frac{\sum_{j=1}^{n} z_{hj}^{(r)} \left[\log(u_{hj}^{(r)}) - u_{hj}^{(r)} - \log\left(\frac{\nu_h^{(r)}+p}{2}\right) + \psi\left(\frac{\nu_h^{(r)}+p}{2}\right)\right]}{\sum_{j=1}^{n} z_{hj}^{(r)}} + \log\left(\frac{\nu_h}{2}\right) - \psi\left(\frac{\nu_h}{2}\right) + 1 = 0, \tag{30}$$

where $\psi(\cdot)$ denotes the Digamma function.

# Appendix B The JCM-MST Model

In this Appendix, we describe the EM algorithm for estimating the parameters of the model described in Section 4 where the components are multivariate skew $t$-densities given by (8).

## B.1 E-step

As in Appendix Appendix A, on the E-step of the EM algorithm we have to calculate the conditional expectation of the complete-data log likelihood given the observed data. It requires the following conditional expectations to be calculated,

$$\begin{aligned} z_{hj}^{(r)} &= E_{\boldsymbol{\Psi}^{(r)}}\left(z_{hj}=1 \mid \boldsymbol{y}_j\right), & (31)\\ u_{hj}^{(r)} &= E_{\boldsymbol{\Psi}^{(r)}}\left(w_{hj} \mid \boldsymbol{y}_j, z_{hj}=1\right), & (32)\\ S_{1,hj}^{(r)} &= E_{\boldsymbol{\Psi}^{(r)}}\left(w_{hj}u_{hj} \mid \boldsymbol{y}_j, u_{hj}>0, z_{hj}=1\right), & (33)\\ S_{2,hj}^{(r)} &= E_{\boldsymbol{\Psi}^{(r)}}\left(w_{hj}u_{hj}^2 \mid \boldsymbol{y}_j, u_{hj}>0, z_{hj}=1\right), & (34)\\ \boldsymbol{S}_{3,hj}^{(r)} &= E_{\boldsymbol{\Psi}^{(r)}}\left(w_{hj}\left(\boldsymbol{a}_{hj}-\mathbf{1}_p\right)\left(\boldsymbol{a}_{hj}-\mathbf{1}_p\right)^T \mid \boldsymbol{y}_j, u_{hj}>0, z_{hj}=1\right), & (35)\\ S_{4,hj}^{(r)} &= E_{\boldsymbol{\Psi}^{(r)}}\left(w_{hj}b_{hj}^2 \mid \boldsymbol{y}_j, u_{hj}>0, z_{hj=1}\right), & (36)\\ \boldsymbol{S}_{5,hj}^{(r)} &= E_{\boldsymbol{\Psi}^{(r)}}\left(w_{hj}u_{hj}(\boldsymbol{y}_j-\boldsymbol{\zeta}_h\boldsymbol{a}_{hj}-\mathbf{1}_p b_{hj}) \mid \boldsymbol{y}_j, u_{hj}>0, z_{hj}=1\right), & (37)\\ \boldsymbol{S}_{6,hj}^{(r)} &= E_{\boldsymbol{\Psi}^{(r)}}\left(w_{hj}\boldsymbol{\epsilon}_{hj}\boldsymbol{\epsilon}_{hj}^T \mid \boldsymbol{y}_j, u_{hj}>0, z_{hj}=1\right), & (38)\\ \boldsymbol{S}_{7,hj}^{(r)} &= E_{\boldsymbol{\Psi}^{(r)}}\left(w_{hj}\boldsymbol{A}_{hj}\boldsymbol{\Sigma}_h^{(r)^{-1}}\boldsymbol{A}_{hj} \mid by_j, u_{hj}>0, z_{hj}=1\right), & (39)\\ \boldsymbol{S}_{8,hj}^{(r)} &= E_{\boldsymbol{\Psi}^{(r)}}\left(w_{hj}\boldsymbol{A}_{hj}\boldsymbol{\Sigma}_h^{(r)^{-1}}(\boldsymbol{y}_j-\mathbf{1}_p b_{hj}-\boldsymbol{\delta}_h u_{hj}) \mid \boldsymbol{y}_j, u_{hj}>0, z_{hj}=1\right), & (40) \end{aligned}$$

where $\boldsymbol{\epsilon}_{hj}=(\boldsymbol{y}_j-\boldsymbol{\zeta}_h\boldsymbol{a}_{hj}-\mathbf{1}_p b_{hj}-\boldsymbol{\delta}_h u_{hj})$. For brevity of notation, we suppress the fact that $z_{hj}=1$ in our subsequent equations.

It is easy to show that

$$z_{hj}^{(r)} = \frac{\pi_h f(\boldsymbol{y}_j; \boldsymbol{\mu}_h^{(r)}, \tilde{\boldsymbol{\Omega}}_h^{(r)}, \boldsymbol{\delta}_h^{(r)}, v_h^{(r)})}{\sum_{h=1}^g \pi_h^{(r)} f(\boldsymbol{y}_j; \boldsymbol{\mu}_h^{(r)}, \tilde{\boldsymbol{\Omega}}_h^{(r)}, \boldsymbol{\delta}_h^{(r)}, \nu_h^{(r)})}, \tag{41}$$

where $f(\boldsymbol{y}_j; \boldsymbol{\mu}_h^{(r)}, \boldsymbol{\Omega}_h^{(r)}, \boldsymbol{\delta}_h^{(r)}, \nu_h^{(r)})$ is the density function pf the skew $t$-distribution, given by

$$2t_p\left(\boldsymbol{y}_j; \boldsymbol{\mu}_h^{(r)}, \tilde{\boldsymbol{\Omega}}_h^{(r)}, \nu_h^{(r)}\right) T_1\left(\frac{\xi_{hj}^{(r)}}{\sigma_h^{(r)}}\sqrt{\frac{\nu_h^{(r)}+p}{\nu_h^{(r)}+d_h^{(r)}(\boldsymbol{y}_j)}}; 0, 1, \nu_h^{(r)}+p\right), \tag{42}$$

and

$$\begin{aligned} \boldsymbol{\zeta}_h^{(r)} &= \mathrm{DIAG}(\boldsymbol{\mu}_h^{(r)}),\\ \boldsymbol{\Omega}_h^{(r)} &= \boldsymbol{\zeta}_h^{(r)}\boldsymbol{\Psi}_h^{(r)}\boldsymbol{\zeta}_h^{(r)^T} + \xi_{2h}^{(r)^2}\mathbf{1}_p\mathbf{1}_p^T + \boldsymbol{\Sigma}_h^{(r)},\\ \tilde{\boldsymbol{\Omega}}_h^{(r)} &= \boldsymbol{\zeta}_h^{(r)}\boldsymbol{\Psi}_h^{(r)}\boldsymbol{\zeta}_h^{(r)^T} + \xi_{2h}^{(r)^2}\mathbf{1}_p\mathbf{1}_p^T + \boldsymbol{\Sigma}_h^{(r)} + \boldsymbol{\delta}_h\boldsymbol{\delta}_h^T,\\ \xi_{hj}^{(r)} &= \boldsymbol{\delta}_h^{(r)^T}\tilde{\boldsymbol{\Omega}}_h^{(r)}\left(\boldsymbol{y}_j-\boldsymbol{\mu}_h^{(r)}\right),\\ \sigma_h^{(r)^2} &= 1-\boldsymbol{\delta}_h^{(r)^T}\tilde{\boldsymbol{\Omega}}_h^{(r)^{-1}}\boldsymbol{\delta}_h^{(r)},\\ d_h^{(r)}(\boldsymbol{y}_j) &= (\boldsymbol{y}_j-\boldsymbol{\mu}_h^{(r)})^T\tilde{\boldsymbol{\Omega}}_h^{(r)^{-1}}(\boldsymbol{y}_j-\boldsymbol{\mu}_h^{(r)}). \end{aligned}$$

Concerning the half-normal distribution, we have that the mean and variance of $u_{hj}$ conditional on $\boldsymbol{y}_j$ and $w_{hj}$, are given by

$$E_{\boldsymbol{\Psi}^{(r)}}(u_{hj} \mid \boldsymbol{y}_j, w_{hj}) = \xi_{hj}^{(r)}$$
$$\text{var}_{\boldsymbol{\Psi}^{(r)}}(u_{hj} \mid \boldsymbol{y}_j, w_{hj}) = \sigma_h^{(r)^2}/w_{hj}.$$

It follows then that we can obtain the following (conditional) moments of the half-normal distribution,

$$\begin{aligned} E_{\boldsymbol{\Psi}^{(r)}}(u_{hj} \mid \boldsymbol{y}_j, w_{hj}, u_{hj} > 0) &= \xi_{hj} + \frac{\sigma_h^{(r)}}{\sqrt{w_{hj}}} \frac{\phi\left(\frac{\xi_{hj}^{(r)}}{\sigma_h^{(r)}}\sqrt{w_{hj}}\right)}{\Phi\left(\frac{\xi_{hj}^{(r)}}{\sigma_h^{(r)}}\sqrt{w_{hj}}\right)}, \\ E_{\boldsymbol{\Psi}^{(r)}}(u_{hj}^2 \mid \boldsymbol{y}_j, w_{hj}, u_{hj} > 0) &= \xi_{hj}^{(r)2} + \frac{\sigma_h^{(r)4}}{w_{hj}} + \frac{\xi_{hj}^{(r)}\sigma_h^{(r)}}{\sqrt{w_{hj}}} \frac{\phi\left(\frac{\xi_{hj}^{(r)}}{\sigma_h^{(r)}}\sqrt{w_{hj}}\right)}{\Phi\left(\frac{\xi_{hj}^{(r)}}{\sigma_h^{(r)}}\sqrt{w_{hj}}\right)}, \end{aligned} \tag{43}$$

where $\phi$ and $\Phi$ denote the standard univariate normal density and (cumulative) distribution function, respectively.

To find the corresponding moments unconditional on $w$, we need the conditional density function of $w$ given $y$,

$$f(w \mid \boldsymbol{y}_j, u_{hj} > 0) = \frac{\Phi\left(\frac{\xi_{hj}^{(r)}}{\sigma_h^{(r)}}\sqrt{w}\right)}{T_1\left(\frac{\xi_{hj}^{(r)}}{\sigma_h^{(r)}}\sqrt{\frac{\nu_h^{(r)}+p}{\nu_h^{(r)}+d_h^{(r)}(\boldsymbol{y}_j)}}; 0, 1, \nu_h^{(r)} + p\right)} f_G\left(w; \frac{\nu_h^{(r)} + p}{2}, \frac{\nu_h^{(r)} + d_h^{(r)}(\boldsymbol{y}_j)}{2}\right), \tag{44}$$

where $f_G(\cdot; \alpha, \beta)$ denotes the gamma density function with shape and scale parameters given by $\alpha$ and $\beta$ respectively.
It follows that

$$\begin{aligned} w_{hj}^{(r)} &= E_{\boldsymbol{\Psi}^{(r)}}(w_{hj} \mid \boldsymbol{y}_j, u_{hj} > 0, z_{hj} = 1) \\ &= \int_0^\infty \frac{\Phi\left(\frac{\xi_{hj}^{(r)}}{\sigma_h^{(r)}}\sqrt{w}\right)}{T_1\left(\frac{\xi_{hj}^{(r)}}{\sigma_h^{(r)}}\sqrt{\frac{\nu_h^{(r)}+p}{\nu_h^{(r)}+d_h^{(r)}(\boldsymbol{y}_j)}};0,1,\nu_h^{(r)}+p\right)} \frac{\left(\frac{\nu_h^{(r)}+d_h^{(r)}(\boldsymbol{y}_j)}{2}\right)^{\frac{\nu_h^{(r)}+p}{2}}}{\Gamma\left(\frac{\nu_h^{(r)}+p}{2}\right)} w^{\frac{\nu_h^{(r)}+p+2}{2}-1} e^{-\frac{w\left(\nu_h^{(r)}+d_h^{(r)}(\boldsymbol{y}_j)\right)}{2}} dw \\ &= \frac{\nu_h^{(r)}+p}{\nu_h^{(r)}+d_h^{(r)}(\boldsymbol{y}_j)} \frac{T_1\left(\frac{\xi_{hj}^{(r)}}{\sigma_h^{(r)}}\sqrt{\frac{\nu_h^{(r)}+p+2}{\nu_h^{(r)}+d_h^{(r)}(\boldsymbol{y}_j)}};0,1,\nu^{(r)}+p+2\right)}{T_1\left(\frac{\xi_{hj}^{(r)}}{\sigma_h^{(r)}}\sqrt{\frac{\nu_h^{(r)}+p}{\nu_h^{(r)}+d_h^{(r)}(\boldsymbol{y}_j)}};0,1,\nu_h^{(r)}+p\right)}, \end{aligned} \tag{45}$$

and

$$\begin{aligned} \boldsymbol{V}_{1,hj}^{(r)} &= E_{\boldsymbol{\Psi}^{(r)}}\left(\sqrt{w_{hj}}\frac{\phi\left(\frac{\xi_{hj^{(r)}}}{\sigma_h^{(r)}}\sqrt{w_{hj}}\right)}{\Phi\left(\frac{\xi_{hj}^{(r)}}{\sigma^{(r)}}\sqrt{w_{hj}}\right)} \mid \boldsymbol{y}_j, u_{hj}>0\right) \\ &= \left[\sqrt{\pi\left(\nu_h^{(r)}+\tilde{d}_h^{(r)}(\boldsymbol{y}_j)\right)}T_1\left(\frac{\xi_{hj}^{(r)}}{\sigma_h^{(r)}}\sqrt{\frac{\nu_h^{(r)}+p}{\nu_h^{(r)}+d_h^{(r)}(\boldsymbol{y}_j)}}\right)\right]^{-1}\left(\frac{\nu_h^{(r)}+d_h^{(r)}(\boldsymbol{y}_j)}{\nu_h+\tilde{d}_h^{(r)}(\boldsymbol{y}_j)}\right)^{\frac{\nu_h^{(r)}+p}{2}}\frac{\Gamma\left(\frac{\nu_h^{(r)}+p+1}{2}\right)}{\Gamma\left(\frac{\nu_h^{(r)}+p}{2}\right)}, \end{aligned}$$

where $\tilde{d}_h^{(r)}(\boldsymbol{y}_j) = d_h^{(r)}(\boldsymbol{y}_j) + \frac{\xi_{hj}^{(r)^2}}{\sigma_h^2}$.

We are now in a position to calculate the required conditional expectations,

$$\begin{aligned} S_{1,hj}^{(r)} &= E_{\boldsymbol{\Psi}^{(r)}}(w_{hj}u_{hj} \mid \boldsymbol{y}_j, u_{hj}>0) \\ &= w_{hj}^{(r)}\xi_{hj}^{(r)} + \sigma_h^{(r)}\boldsymbol{V}_{1,hj}^{(r)}, \quad (46) \\ S_{2,hj}^{(r)} &= E_{\boldsymbol{\Psi}^{(r)}}\left(w_{hj}u_{hj}^2 \mid \boldsymbol{y}_j, u_{hj}>0\right), \\ &= w_{hj}^{(r)}\xi_{hj}^{(r)^2} + \sigma_h^{(r)^4} + \sigma_h^{(r)}\xi_{hj}^{(r)}V_{1,hj}^{(r)}. \quad (47) \end{aligned}$$

To calculate the remaining conditional expectations, note that

$$\begin{aligned} &E_{\boldsymbol{\Psi}^{(r)}}\left(\begin{bmatrix}\boldsymbol{a}_{hj}\\ b_{hj}\end{bmatrix} \mid \boldsymbol{y}_j, w_{hj}, u_{hj}\right) \\ &= \begin{bmatrix}\boldsymbol{1}_p\\ 0\end{bmatrix} + \begin{bmatrix}\boldsymbol{\Psi}_h^{(r)}\boldsymbol{\zeta}_h^{(r)}\\ \xi_{2h}^{(r)^2}\boldsymbol{1}_p^T\end{bmatrix}\boldsymbol{\Omega}_h^{(r)^{-1}}\left(\boldsymbol{y}_j-\boldsymbol{\mu}_h^{(r)}-\boldsymbol{\delta}_h^{(r)}u_{hj}\right) \quad (48) \end{aligned}$$

and

$$\begin{aligned} \boldsymbol{V}_{2,hj}^{(r)} &= \mathrm{cov}_{\boldsymbol{\Psi}^{(r)}}\left(\begin{bmatrix}\boldsymbol{a}_{hj}\\ b_{hj}\end{bmatrix} \mid \boldsymbol{y}_j, w_{hj}, u_{hj}\right) \\ &= \frac{1}{w_{hj}}\begin{bmatrix}\boldsymbol{\Psi}_h^{(r)} & 0\\ 0 & \xi_{2h}^{(r)^2}\end{bmatrix} - \frac{1}{w_{hj}}\begin{bmatrix}\boldsymbol{\Psi}_h^{(r)}\boldsymbol{\zeta}_h^{(r)}\\ \xi_{2h}^{(r)^2}\boldsymbol{1}_p^T\end{bmatrix}\boldsymbol{\Omega}_h^{(r)^{-1}}\begin{bmatrix}\boldsymbol{\zeta}_h^{(r)}\boldsymbol{\Psi}_h^{(r)} & \xi_{2h}^{(r)^2}\boldsymbol{1}_p\end{bmatrix}. \quad (49) \end{aligned}$$

From the above, we have

$$\begin{aligned} E_{\boldsymbol{\Psi}^{(r)}}\left(\boldsymbol{a}_{hj}-\boldsymbol{1}_p \mid \boldsymbol{y}_j, w_{hj}, u_{hj}\right) &= \boldsymbol{\Psi}_h^{(r)}\boldsymbol{\zeta}^{(r)}\boldsymbol{\Omega}_h^{(r)^{-1}}\left(\boldsymbol{y}_j-\boldsymbol{\mu}_h^{(r)}-\boldsymbol{\delta}_h^{(r)}u_{hj}\right), \\ E_{\boldsymbol{\Psi}^{(r)}}\left(b_{hj} \mid \boldsymbol{y}_j, w_{hj}, u_{hj}\right) &= \xi_{2h}^{(r)^2}\boldsymbol{1}_p^{(r)^T}\boldsymbol{\Omega}_h^{(r)^{-1}}\left(\boldsymbol{y}_j-\boldsymbol{\mu}_h^{(r)}-\boldsymbol{\delta}_h^{(r)}u_{hj}\right), \end{aligned}$$

and

$$\begin{aligned} \mathrm{cov}_{\boldsymbol{\Psi}^{(r)}}\left(\boldsymbol{a}_{hj}-\boldsymbol{1}_p \mid \boldsymbol{y}_j, w_{hj}, u_{hj}\right) &= \frac{1}{w_{hj}}\left(\boldsymbol{\Psi}_h^{(r)}-\boldsymbol{\Psi}_h^{(r)}\boldsymbol{\zeta}_h^{(r)^T}\boldsymbol{\Omega}_h^{(r)^{-1}}\boldsymbol{\zeta}_h^{(r)}\boldsymbol{\Psi}_h^{(r)}\right), \\ \mathrm{cov}_{\boldsymbol{\Psi}^{(r)}}\left(b_{hj} \mid \boldsymbol{y}_j, w_{hj}, u_{hj}\right) &= \frac{1}{w_{hj}}\left(\xi_{2h}^{(r)^2}-\xi_{2h}^{(r)^2}\boldsymbol{1}_p^T\boldsymbol{\Omega}_h^{(r)^{-1}}\boldsymbol{1}_p\right), \\ \mathrm{cov}_{\boldsymbol{\Psi}^{(r)}}\left(\boldsymbol{a}_{hj}b_{hj} \mid \boldsymbol{y}_j, w_{hj}, u_{hj}\right) &= \frac{1}{w_{hj}}\left(-\xi_{2h}^{(r)^2}\boldsymbol{\Psi}_h^{(r)}\boldsymbol{\zeta}_h^{(r)}\boldsymbol{\Omega}_h^{(r)^{-1}}\boldsymbol{1}_p\right). \end{aligned}$$

Also,

$$\operatorname{cov}_{\boldsymbol{\Psi}^{(r)}}\left(\boldsymbol{y}_j-\boldsymbol{\zeta}_h^{(r)}\boldsymbol{a}_{hj}-\mathbf{1}_p b_{hj}\mid \boldsymbol{y}_j,w_{hj},u_{hj}\right)$$
$$=\left[\boldsymbol{\zeta}_h^{(r)}\ \ \mathbf{1}_p\right]\boldsymbol{V}_{2,hj}^{(r)}\begin{bmatrix}\boldsymbol{\zeta}_h^{(r)^T}\\ \mathbf{1}_p^T\end{bmatrix}+\boldsymbol{V}_{3,hj}^{(r)}\boldsymbol{V}_{3,hj}^{(r)^T}$$
$$=\frac{1}{w_{hj}}\left(\boldsymbol{V}_{4,hj}^{(r)}-\boldsymbol{V}_{4,hj}^{(r)}\boldsymbol{\Omega}_h^{(r)^{-1}}\boldsymbol{V}_{4,hj}^{(r)}\right)+\boldsymbol{V}_{3,hj}^{(r)}\boldsymbol{V}_{3,hj}^{(r)^T},$$

where

$$\boldsymbol{V}_{3,hj}^{(r)}=\boldsymbol{y}_j-\boldsymbol{\mu}_h^{(r)}-\left(\boldsymbol{\zeta}_h^{(r)}\boldsymbol{\Psi}_h^{(r)}+\xi_h^{(r)^2}\mathbf{1}_p\right)\boldsymbol{\Omega}_h^{(r)^{-1}}\left(\boldsymbol{y}_j-\boldsymbol{\mu}_h^{(r)}\right)$$

and

$$\boldsymbol{V}_{4,hj}^{(r)}=\boldsymbol{\zeta}_h^{(r)}\boldsymbol{\Psi}_h\boldsymbol{\zeta}_h^{(r)}+\xi_{2h}^{(r)^2}\mathbf{1}_p\mathbf{1}_p^T.$$

It follows that $\boldsymbol{S}_{3,hj}^{(r)}$ and $S_{4,hj}^{(r)}$ are given by

$$\begin{aligned}\boldsymbol{S}_{3,hj}^{(r)} \;=\; & \left(\boldsymbol{\Psi}_h^{(r)}-\boldsymbol{\Psi}_h^{(r)}\boldsymbol{\zeta}_h^{(r)^T}\boldsymbol{\Omega}_h^{(r)^{-1}}\boldsymbol{\zeta}_h^{(r)}\boldsymbol{\Psi}_h^{(r)}\right)\\ & + w_{hj}^{(r)}\boldsymbol{\Psi}_{hj}^{(r)}\boldsymbol{\zeta}_h^{(r)^T}\boldsymbol{\Omega}_h^{(r)^{-1}}(\boldsymbol{y}_j-\boldsymbol{\mu}_h^{(r)})(\boldsymbol{y}_j-\boldsymbol{\mu}_h^{(r)})^T\boldsymbol{\Omega}_h^{(r)^{-1}}\boldsymbol{\zeta}_h^{(r)}\boldsymbol{\Psi}_h^{(r)}\\ & + \boldsymbol{\Psi}_h^{(r)}\boldsymbol{\zeta}_h^{(r)^T}\boldsymbol{\Omega}_h^{(r)^{-1}}\boldsymbol{\delta}_h^{(r)}\boldsymbol{\delta}_h^{(r)^T}\boldsymbol{\Omega}_h^{(r)^{-1}}\boldsymbol{\zeta}_h^{(r)}\boldsymbol{\Psi}_h^{(r)}\boldsymbol{S}_{2,hj}^{(r)}\\ & - \boldsymbol{\Psi}_h^{(r)}\boldsymbol{\zeta}_h^{(r)}\boldsymbol{\Omega}_h^{(r)^{-1}}(\boldsymbol{y}_j-\boldsymbol{\mu}_h^{(r)})\boldsymbol{\delta}_h^{(r)T}\boldsymbol{\Omega}_h^{(r)^{-1}}\boldsymbol{\zeta}_h^{(r)}\boldsymbol{\Psi}_h^{(r)}\boldsymbol{S}_{1,hj}^{(r)}\\ & - \boldsymbol{\Psi}_h^{(r)}\boldsymbol{\zeta}_h^{(r)^T}\boldsymbol{\Omega}_h^{(r)^{-1}}\boldsymbol{\delta}_h^{(r)}(\boldsymbol{y}_j-\boldsymbol{\mu}_h^{(r)})^T\boldsymbol{\Omega}_h^{(r)^{-1}}\boldsymbol{\zeta}_h^{(r)}\boldsymbol{\Psi}_h^{(r)}\boldsymbol{S}_{1,hj}^{(r)}\end{aligned}\tag{50}$$

and

$$\begin{aligned}\boldsymbol{S}_{4,hj}^{(r)} \;=\; & \left(\xi_{2h}^{(r)^2}-\xi_{2h}^{(r)^2}\mathbf{1}_p^T\boldsymbol{\Omega}_h^{(r)^{-1}}\mathbf{1}_p\right)\\ & + w_{hj}^{(r)}\xi_{2h}^{(r)^2}\mathbf{1}_p^{(r)^T}\boldsymbol{\Omega}_h^{(r)^{-1}}(\boldsymbol{y}_j-\boldsymbol{\mu}_h^{(r)})(\boldsymbol{y}_j-\boldsymbol{\mu}_h^{(r)})^T\boldsymbol{\Omega}_h^{(r)^{-1}}\mathbf{1}_p\\ & + \xi_{2h}^{(r)^2}\mathbf{1}_p^T\boldsymbol{\Omega}_h^{(r)^{-1}}\boldsymbol{\delta}_h^{(r)}\boldsymbol{\delta}_h^{(r)T}\boldsymbol{\Omega}_h^{(r)^{-1}}\mathbf{1}_p\boldsymbol{S}_{2,hj}^{(r)}\\ & - \xi_h^{(r)^2}\mathbf{1}_p^T\boldsymbol{\Omega}_h^{(r)^{-1}}\boldsymbol{\delta}_h^{(r)}(\boldsymbol{y}_j-\boldsymbol{\mu}_h^{(r)})^T\boldsymbol{\Omega}_h^{(r)^{-1}}\mathbf{1}_p\boldsymbol{S}_{1,hj}^{(r)}\\ & -\xi_h^{(r)^2}\mathbf{1}_p^T\boldsymbol{\Omega}_h^{(r)^{-1}}(\boldsymbol{y}_j-\boldsymbol{\mu}_h^{(r)})\boldsymbol{\delta}_h^{(r)^T}\boldsymbol{\Omega}_h^{(r)^{-1}}\mathbf{1}_p\boldsymbol{S}_{1,hj}^{(r)}.\end{aligned}\tag{51}$$

Further, we have that

$$\begin{aligned}\boldsymbol{S}_{5,hj}^{(r)}=&\left[\boldsymbol{y}_j-\boldsymbol{\zeta}_h^{(r)}\mathbf{1}_p-\left(\boldsymbol{\zeta}_h^{(r)}\boldsymbol{\Psi}_h^{(r)}\boldsymbol{\zeta}_h^{(r)^T}+\xi_h^{(r)^2}\mathbf{1}_p\mathbf{1}_p^T\right)\boldsymbol{\Omega}_h^{(r)^{-1}}\left(\boldsymbol{y}_j-\boldsymbol{\mu}_h^{(r)}\right)\right]S_{1,hj}^{(r)}\\ &+\left(\boldsymbol{\zeta}_h^{(r)}\boldsymbol{\Psi}_h^{(r)}\boldsymbol{\zeta}_h^{(r)^T}+\xi_{2h}^{(r)^2}\mathbf{1}_p\mathbf{1}_p^T\right)\boldsymbol{\Omega}_h^{(r)^{-1}}\boldsymbol{\delta}_h^{(r)}S_{2,hj}^{(r)}\end{aligned}\tag{52}$$

(53)

and

$$\begin{aligned}\boldsymbol{S}_{6,hj}^{(r)}=&\boldsymbol{V}_{4,hj}^{(r)}-\boldsymbol{V}_{4,hj}^{(r)}\boldsymbol{\Omega}_h^{(r)^{-1}}\boldsymbol{V}_{4,hj}^{(r)}+w_{hj}^{(r)}\boldsymbol{V}_{3,hj}^{(r)}\boldsymbol{V}_{3,hj}^{(r)^T}\\ &-\boldsymbol{S}_{5,hj}^{(r)}\boldsymbol{\delta}_h^{(r)^T}-\boldsymbol{\delta}_h^{(r)}\boldsymbol{S}_{5,hj}^{(r)^T}+S_{2,hj}^{(r)}\boldsymbol{\delta}_h^{(r)}\boldsymbol{\delta}_h^{(r)^T}.\end{aligned}\tag{54}$$

To evaluate $\boldsymbol{S}_{7,hj}^{(r)}$, let $\boldsymbol{A}_{hj}$ be the diagonal matrix with $\boldsymbol{a}_{hj}$ as its diagonal elements. Then

$$E_{\boldsymbol{\Psi}^{(r)}}\left(\boldsymbol{A}_{hj}\boldsymbol{\Sigma}_h^{(r)^{-1}}\boldsymbol{A}_{hj} \mid \boldsymbol{y}_j, w_{hj}, u_{hj}\right) = E_{\boldsymbol{\Psi}^{(r)}}\left(\boldsymbol{a}_{hj}\boldsymbol{a}_{hj}^T \mid \boldsymbol{y}_j, w_{hj}, u_{hj}\right) \odot \boldsymbol{\Sigma}_h^{(r)^{-1}}. \tag{55}$$

It is straightforward to show that

$$\begin{aligned}
\boldsymbol{S}_{7,hj}^{(r)} &= \Big\{\left(\boldsymbol{\Psi}_h^{(r)} - \boldsymbol{\Psi}_{ah}^{(r)}\boldsymbol{\zeta}_h^{(r)^T}\boldsymbol{\Omega}_h^{(r)^{-1}}\boldsymbol{\zeta}_h^{(r)}\boldsymbol{\Psi}_h^{(r)}\right) \\
&+ w_{hj}^{(r)}\left[\mathbf{1}_p + \boldsymbol{\Psi}_h^{(r)}\boldsymbol{\zeta}_h^{(r)^T}\boldsymbol{\Omega}_h^{(r)^{-1}}\left(\boldsymbol{y}_j - \boldsymbol{\mu}_h^{(r)}\right)\right]\left[\mathbf{1}_p + \boldsymbol{\Psi}_h^{(r)}\boldsymbol{\zeta}_h^{(r)^T}\boldsymbol{\Omega}_h^{(r)^{-1}}\left(\boldsymbol{y}_j - \boldsymbol{\mu}_h^{(r)}\right)\right]^T \\
&- S_{1,hj}^{(r)}\left[\boldsymbol{\Psi}_h^{(r)}\boldsymbol{\zeta}_h^{(r)^T}\boldsymbol{\Omega}_h^{(r)}\boldsymbol{\delta}_h^{(r)}\right]\left[\mathbf{1}_p + \boldsymbol{\Psi}_h^{(r)}\boldsymbol{\zeta}_h^{(r)^T}\boldsymbol{\Omega}_h^{(r)^{-1}}\left(\boldsymbol{y}_j - \boldsymbol{\mu}_h^{(r)}\right)\right]^T \\
&- S_{1,hj}^{(r)}\left[\mathbf{1}_p + \boldsymbol{\Psi}_h^{(r)}\boldsymbol{\zeta}_h^{(r)^T}\boldsymbol{\Omega}_h^{(r)^{-1}}\left(\boldsymbol{y}_j - \boldsymbol{\mu}_h^{(r)}\right)\right]\left[\boldsymbol{\delta}_h^{(r)^T}\boldsymbol{\Omega}_h^{(r)^{-1}}\boldsymbol{\zeta}_h^{(r)}\boldsymbol{\Psi}_h^{(r)}\right] \\
&+ S_{2,hj}^{(r)}\boldsymbol{\Psi}_h^{(r)}\boldsymbol{\zeta}_h^{(r)^T}\boldsymbol{\Omega}_h^{(r)^{-1}}\boldsymbol{\delta}_h^{(r)}\boldsymbol{\delta}_h^{(r)^T}\boldsymbol{\Omega}_h^{(r)^{-1}}\boldsymbol{\zeta}_h^{(r)}\boldsymbol{\Psi}_h^{(r)}\Big\} \odot \boldsymbol{\Sigma}_h^{(r)^{-1}}.
\end{aligned} \tag{56}$$

To obtain $\boldsymbol{S}_{8,hj}^{(r)}$, observe that

$$\begin{aligned}
& E_{\boldsymbol{\Psi}^{(r)}}\left(\boldsymbol{A}_{hj}\boldsymbol{\Sigma}_h^{(r)^{-1}}\left(\boldsymbol{y}_j - \mathbf{1}_p b_{hj} - \boldsymbol{\delta}_h^{(r)}u_{hj}\right) \mid \boldsymbol{y}_j, w_{hj}, u_{hj}\right) \\
= \; & E_{\boldsymbol{\Psi}^{(r)}}\left(\boldsymbol{A}_{hj} \mid \boldsymbol{y}_j, w_{hj}, u_{hj}\right)\boldsymbol{\Sigma}_h^{(r)^{-1}}\left(\boldsymbol{y}_j - \boldsymbol{\delta}_h^{(r)}u_{hj}\right) - E_{\boldsymbol{\Psi}^{(r)}}\left(\boldsymbol{A}_{hj}b_{hj} \mid \boldsymbol{y}_j, w_{hj}, u_{hj}\right)\boldsymbol{\Sigma}_h^{(r)^{-1}}\mathbf{1}_p.
\end{aligned}$$

It follows that

$$\begin{aligned}
\boldsymbol{S}_{8,hj}^{(r)} &= \mathrm{DIAG}\left[\mathbf{1}_p + \boldsymbol{\Psi}_h^{(r)}\boldsymbol{\zeta}_h^{(r)^T}\boldsymbol{\Omega}_h^{(r)^{-1}}\left(\boldsymbol{y}_j - \boldsymbol{\mu}_h\right)\right]\boldsymbol{\Sigma}_h^{(r)^{-1}}\left(\boldsymbol{y}_j w_{hj}^{(r)} - S_{1,hj}^{(r)}\boldsymbol{\delta}_h^{(r)}\right) \\
&\quad - \mathrm{DIAG}\left(\boldsymbol{\Psi}_h^{(r)}\boldsymbol{\zeta}_h^{(r)^T}\boldsymbol{\Omega}_h^{(r)}\boldsymbol{\delta}_h^{(r)}\right)\boldsymbol{\Sigma}_h^{(r)^{-1}}\left(S_{1,hj}^{(r)}\boldsymbol{y}_j - S_{2,hj}^{(r)}\boldsymbol{\delta}_h^{(r)}\right) \\
&\quad + \mathrm{DIAG}\left(\boldsymbol{V}_{5,hj}^{(r)}\right)\boldsymbol{\Sigma}_h^{(r)^{-1}}\mathbf{1}_p, 
\end{aligned} \tag{57}$$

$$\begin{aligned}
\boldsymbol{V}_{5,hj}^{(r)} &= \xi_{2h}^{(r)^2}\boldsymbol{\Psi}_h^{(r)}\boldsymbol{\zeta}_h^{(r)}\boldsymbol{\Omega}_h^{(r)^{-1}}\mathbf{1}_p \\
&\quad + w_{hj}^{(r)}\xi_{2h}^{(r)^2}\left[\mathbf{1}_p + \boldsymbol{\Psi}_h^{(r)}\boldsymbol{\zeta}_h^{(r)}\boldsymbol{\Omega}_h^{(r)^{-1}}\left(\boldsymbol{y}_j - \boldsymbol{\mu}_h^{(r)}\right)\right]\left(\boldsymbol{y}_j - \boldsymbol{\mu}_h^{(r)}\right)^T\boldsymbol{\Omega}_h^{(r)^{-1}}\mathbf{1}_p \\
&\quad - \boldsymbol{S}_{1,hj}^{(r)}\xi_{2h}^{(r)^2}\left[\mathbf{1}_p + \boldsymbol{\Psi}_h^{(r)}\boldsymbol{\zeta}_h^{(r)^T}\boldsymbol{\Omega}_h^{(r)^{-1}}\left(\boldsymbol{y}_j - \boldsymbol{\mu}_h^{(r)}\right)\right]\boldsymbol{\delta}_h^{(r)}\boldsymbol{\Omega}_h^{(r)^{-1}}\mathbf{1}_p \\
&\quad - \boldsymbol{S}_{1,hj}^{(r)}\xi_{2h}^{(r)^2}\mathbf{1}_p^T\boldsymbol{\Omega}_h^{(r)^{-1}}\boldsymbol{\delta}_h^{(r)T}\left[\left(\boldsymbol{y}_j - \boldsymbol{\mu}_h^{(r)}\right)^T\boldsymbol{\Omega}_h^{(r)^{-1}}\boldsymbol{\zeta}_h^{(r)}\boldsymbol{\Psi}_h^{(r)} + \mathbf{1}_p^T\right] \\
&\quad + \boldsymbol{S}_{2,hj}^{(r)}\xi_{2h}^{(r)^2}\mathbf{1}_p^T\boldsymbol{\Omega}_h^{(r)^{-1}}\boldsymbol{\delta}_h^{(r)}\boldsymbol{\delta}_h^{(r)T}\boldsymbol{\Omega}_h^{(r)^{-1}}\mathbf{1}_p.
\end{aligned} \tag{58}$$

## B.2 M-step

The estimates of the parameters are updated on the M-step by maximizing the $Q$-function over the parameter space. It follows that

$$\pi_h^{(r+1)} = \frac{1}{n} \sum_{j=1^n} z_{hj}^{(r)}, \tag{59}$$

$$\boldsymbol{\Psi}_h^{(r+1)} = \frac{\sum_{j=1}^n z_{hj}^{(r)} \boldsymbol{S}_{3,hj}^{(r)}}{\sum_{j=1}^n z_{hj}^{(r)}}, \tag{60}$$

$$\xi_{2h}^{(r+1)^2} = \frac{\sum_{j=1}^n z_{hj}^{(r)} S_{4,hj}^{(r)}}{\sum_{j=1}^n z_{hj}^{(r)}}, \tag{61}$$

$$\boldsymbol{\delta}_h^{(r+1)} = \frac{\sum_{j=1}^n z_{hj}^{(r)} \boldsymbol{S}_{5,hj}^{(r)}}{\sum_{j=1}^n z_{hj}^{(r)} S_{2,hj}^{(r)}}, \tag{62}$$

$$\boldsymbol{\Sigma}_h^{(r+1)} = \frac{\sum_{j=1}^n z_{hj}^{(r)} \boldsymbol{S}_{6,hj}^{(r)}}{\sum_{j=1}^n z_{hj}^{(r)}}, \tag{63}$$

and

$$\boldsymbol{\mu}_h^{(r+1)} = \left( \sum_{j=1}^n z_{hj}^{(r)} \boldsymbol{S}_{7,hj}^{(r)} \right)^{-1} \left( \sum_{j=1}^n z_{hj}^{(r)} \boldsymbol{S}_{8,hj}^{(r)} \right). \tag{64}$$

The update of the degrees of freedom $\nu_h^{(r+1)}$ is given implicitly as a solution of the equation,

$$\frac{\sum_{j=1}^n z_{hj}^{(r)} \left[ \psi\left( \frac{\nu_h^{(r)}+p}{2} \right) - \log\left( \frac{\nu_h^{(r)}+d_h^{(r)}(\boldsymbol{y}_j)}{2} \right) - \frac{\nu_h^{(r)}+p}{\nu_h^{(r)}+d_h^{(r)}(\boldsymbol{y}_j)} \right]}{\sum_{j=1}^n z_{hj}^{(r)}} + \log\left(\frac{\nu_h}{2}\right) - \psi\left(\frac{\nu_h}{2}\right) + 1 = 0. \tag{65}$$

where $\psi(\cdot)$ denotes the Digamma function.